# A Study on Overfitting in Deep Reinforcement Learning

Chiyuan Zhang chiyuan@google.com
Oriol Vinyals vinyals@google.com
Remi Munos munos@google.com
Samy Bengio bengio@google.com

**Abstract**

Recent years have witnessed significant progresses in deep Reinforcement Learning (RL). Empowered with large scale neural networks, carefully designed architectures, novel training algorithms and massively parallel computing devices, researchers are able to attack many challenging RL problems. However, in machine learning, more training power comes with a potential risk of more overfitting. As deep RL techniques are being applied to critical problems such as healthcare and finance, it is important to understand the generalization behaviors of the trained agents. In this paper, we conduct a systematic study of standard RL agents and find that they could overfit in various ways. Moreover, overfitting could happen "robustly": commonly used techniques in RL that add stochasticity do not necessarily *prevent* or *detect* overfitting. In particular, the same agents and learning algorithms could have drastically different test performance, even when all of them achieve optimal rewards during training. The observations call for more principled and careful evaluation protocols in RL. We conclude with a general discussion on overfitting in RL and a study of the generalization behaviors from the perspective of inductive bias.

## 1 Introduction

Deep neural networks have proved to be effective function approximators in Reinforcement Learning (RL). Significant progress is seen in many RL problems ranging from board games like Go (Silver et al., 2016, 2017b), Chess and Shogi (Silver et al., 2017a), video games like Atari (Mnih et al., 2015) and StarCraft (Vinyals et al., 2017), to real world robotics and control tasks (Lillicrap et al., 2016). Most of these successes are due to improved training algorithms, carefully designed neural network architectures and powerful hardware. For example, in AlphaZero (Silver et al., 2017a), 5,000 1st-generation TPUs and 64 2nd-generation TPUs are used during self-play based training of agents with deep residual networks (He et al., 2016).

On the other hand, learning with high-capacity models and long stretched training time on powerful devices could lead to potential risk of *overfitting* (Hardt et al., 2016; Lin et al., 2016). As a fundamental trade-off in machine learning, preventing overfitting by properly controlling or regularizing the training is key to out-of-sample *generalization*. Studies of overfitting could be performed from the theory side, where generalization guarantees are derived for specific learning algorithms; or from the practice side, where carefully designed experimental protocols like *cross validation* are used as proxy to certify the generalization performance.

Unfortunately, in the regime of deep RL, systematic studies of generalization behaviors from either theoretical or empirical perspectives are falling behind the rapid progresses from the algorithm development and application side. The current situation not only makes it difficult to understand the test behaviors like the vulnerabilities to potential adversarial attacks (Huang et al., 2017), but also renders some results difficult to reproduce or compare (Henderson et al., 2017; Machado et al., 2017). Meanwhile, as RL techniques being applied to important fields concerning our everyday life such as healthcare (Chakraborty & Murphy, 2014; Kosorok & Moodie, 2015) and finance (Li et al., 2009; Yu et al., 2009; Deng et al., 2017), and fundamental infrastructures such as power grids (Wen et al., 2015; Glavic et al., 2017) and traffic control (Mannion et al., 2016; van der Pol & Oliehoek, 2016), it is becoming crucial to establish good understandings of the representation learning, long term planning, exploration and adaptation abilities of deep RL agents before deploying them into real world applications.

Table 1: Techniques to inject stochasticity in Atari evaluations.

| Technique | References |
|---|---|
| Stochastic policy | Hausknecht & Stone (2015) |
| Random starts | Mnih et al. (2015); Hausknecht & Stone (2015); Nair et al. (2015) |
| Sticky actions | Hausknecht & Stone (2015); Brockman et al. (2016); Machado et al. (2017) |
| Frame skipping | Brockman et al. (2016) |

The formal setup of RL problems are usually in the "continual learning" setting without explicitly separated *training* and *testing* stages. The goal is to achieve the highest cumulative reward, typically discounted over time. In many applications such as game tournaments or robotics, agents are usually deployed to the "test" environments after training. In this case, the performance on the training set might not reflect the true performance on the unseen (but statistically similar) test data. Unlike in the supervised learning case, there are no standard experimental protocols or performance measures yet even on popular tasks like Atari (Machado et al., 2017). Since an isolated test set is *not* used, various techniques are proposed (see Table 1) to inject stochasticity to the environments during testing to avoid artificially high performance by simple algorithms like *trajectory tree* (Kearns et al., 2000) or *Brute* (Machado et al., 2017) that optimize over open-loop sequences of actions without even looking at the states. However, very few formal studies are found on whether those techniques are effective at *preventing* or even *detecting* overfitting in general.

In this paper, we conduct a systematic study of the overfitting and generalization behavior of standard deep RL agents, using a highly configurable maze environment that generates games with various difficulties and regularities. The main contributions of this work are the following:

1. Systematic study of the generalization and memorization capacities of deep RL agents under regular and randomized maze games. We found that the agents are able to memorize a large collection of (even random) training mazes. With the same (optimal) training performances, the test performance could vary drastically.

2. Evaluations of standard techniques used in the literature for avoiding overfitting in our framework. We found that deep RL agents could overfit "robustly", and added stochasticity to the environments does not necessarily *prevent* or *detect* overfitting.

3. An interpretation of the drastically different generalization behaviors under different environments from the perspective of inductive bias and empirical studies to verify the hypothesis.

## 2 Related Work

Characterization of generalization properties has been a prevalent topic in theoretical RL studies. Algorithms and regret analysis for stochastic (Auer et al., 2002) and adversarial bandits (Auer, 2000; Bubeck & Cesa-Bianchi, 2012), for contextual bandits (Agarwal et al., 2014), contextual decision processes (Jiang et al., 2017) and RL (Jaksch et al., 2010; Azar et al., 2017), as well as *Probably Approximately Correct* (PAC) (Strehl et al., 2009; Lattimore & Hutter, 2014; Dann et al., 2017) guarantees are studied in the literature. However, most of the theoretical analysis focused on simple settings with small discrete state and action spaces, or more restricted cases like (possibly contextual) *multi-arm bandit* problems. The algorithmic complexities and generalization bounds typically scale polynomially with the cardinality of the state and action spaces, and therefore not easily applicable to many real world problems.

In the regime of large-scale discrete RL or continuous control problems, many empirical successes are achieved with algorithms using deep neural networks as function approximators. But the theoretical understandings for those scenarios are much less developed. Furthermore, even the empirical results reported in the literature are sometimes difficult to compare due to non-standard evaluation protocols and poor reproducibility (Henderson et al., 2017), which has raised a lot of concerns in the community recently.

Some previous efforts were made towards more standard evaluation for RL algorithms. Nouri et al. (2009) proposed benchmark data repositories, similar to the UCI repository for supervised learning (Lichman, 2013), to evaluate the generalization performance of Q-function estimation (Sutton & Barto, 1998) of a fixed policy via separate training and testing trajectories. However, the proposed framework ignored some important components in RL like action selection and exploration. In Whiteson et al. (2011), a "multiple environments" paradigm is proposed to separate the training and testing stages. In Machado et al. (2017), with the introduction of a new version of the *Arcade Learning Environment* for Atari 2600 games, it summarized the diversity in the evaluation methodologies used by the research community, and proposed to use *sticky actions* to detect agent overfitting by randomly repeating the previous action chosen by the agents during the evaluation time.

Our methodology is inspired by Zhang et al. (2017), which studied the capacity of large neural networks in supervised learning via randomization tests. In this paper, we extend the study to the setting of RL and evaluate the capability of memorizing random tasks and study the phenomenon of overfitting in deep RL. Raghu et al. (2017) is a closely related work that studied RL under a set of carefully controlled combinatorial games. While their primary concern is the power of various RL learning algorithms, we mainly focus on the topic of generalization and overfitting in RL.

# 3 Generalization in Deep RL Agents

In this section, we study the generalization performance of deep RL agents on a gridworld maze environment. Formally, we denote an RL task by $(\mathcal{M} = (\mathcal{S}, \mathcal{A}, \mathcal{P}, r), \mathcal{P}_0)$, where $\mathcal{M}$ is a Markov Decision Process (MDP) with the state space $\mathcal{S}$, the action space $\mathcal{A}$, the state transition probability kernel $\mathcal{P}$ and the immediate reward function $r$. In addition, $\mathcal{P}_0$ is a probability distribution on the initial states $\mathcal{S}_0 \subset \mathcal{S}$. We use an episodic setting, where each episode starts with a state independently sampled from $\mathcal{P}_0$, and ends in finite $T \leq T_{\text{max}}$ steps. Our performance evaluation metric is the episode reward, which is non-discounted sum of all rewards collected in a full episode. We are interested in the expected performance with respect to all the randomness including the initial state sampling. In practice, $\mathcal{P}_0$ is either unknown or difficult to enumerate, but we have access to a set of i.i.d. samples $\hat{\mathcal{S}}_0$ from it. We split $\hat{\mathcal{S}}_0$ into disjoint *training* and *test* sets. During training, the agents have access to initial states only from the training set. The test performance is calculated based on only the initial states from the test set. The generalization performance is the difference between the test and training performances.

## 3.1 Task Setup and Evaluation Protocol

The RL environment used in this study is a gridworld maze, in which an agent seeks objects with positive rewards while avoiding ones with negative rewards. We use three different variants with increasing difficulty: `BASIC`, `BLOCKS` and `TUNNEL`. For ease of comparison, the maximum achievable reward per episode is fixed at 2.1 for every maze. Please refer to appendix A.1 for details and Figure 1a for an example of a `TUNNEL` maze. The reason to use a customized environment instead of more popular platforms like Atari is to have precise controls and customization over the tasks for comparative studies. The goal of this paper is neither to propose novel RL algorithms nor to demonstrate the power of RL on new challenging tasks, but to compare and analyze the behaviors of deep RL agents under controlled experiments.

*Asynchronous Actor-Critic Agent* (A3C; Mnih et al., 2016) is used for learning, where each worker interacts with an independent copy of the game environment, and asynchronously updates the neural network weights in

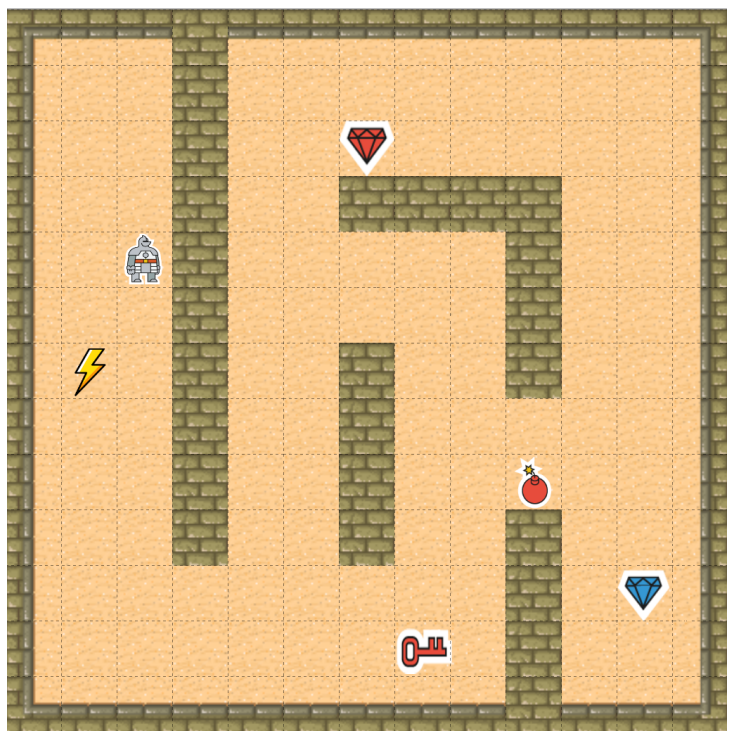

(a) An example of a TUNNEL maze.

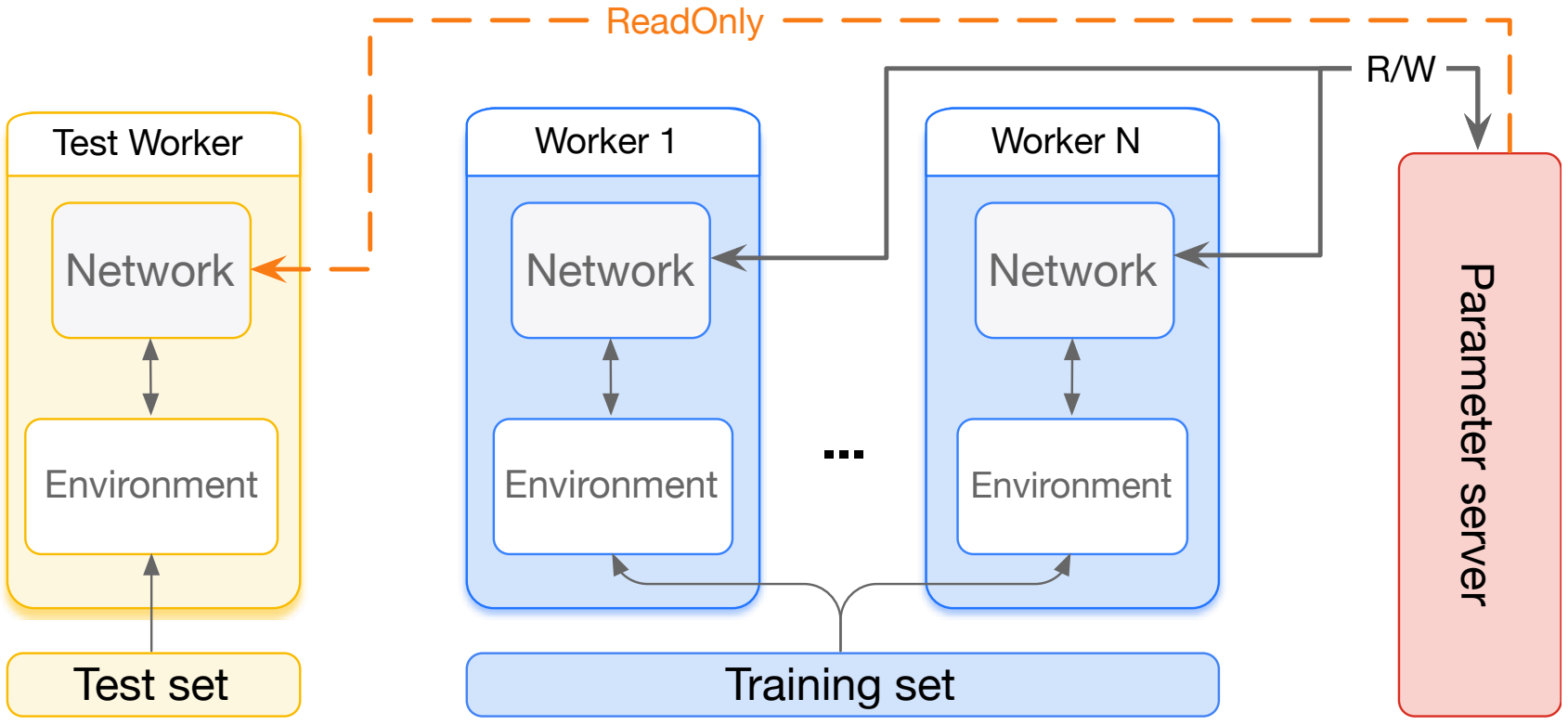

(b) Parallel training and evaluating workers, with separate training and test environments.

Figure 1: Illustration of the maze game and the asynchronous training / evaluation protocol with parameter servers.

the *parameter server*. Because the games have finite number of steps, we follow an episodic setting. Each time an episode ends, the environment is reset and reinitialized with a new random maze configuration. To mimic the behavior of video games, we also introduce the notion of *levels*. The level id is a random seed that determines the initial state for the corresponding level. Furthermore, as video games normally contain a finite number of levels, we sample a pool of ids for the training levels, and a *separate* pool for testing levels.

An important modification we make to the A3C framework is to include a *test worker*. During training, it continuously pulls the latest network weights from the parameter server to compute policies to interact with the environments on the isolated *test levels*. But it does *not* compute or send back gradients. Figure 1b illustrates the protocol. With this setup, we are able to create the learning and testing curves as commonly seen in the supervised learning setting.

### 3.2 Training Optimality and Generalization

In this subsection, we study the generalization performance of trained agents along three axes — the training time, the training set size, and the difficulty of the tasks. Specifically, we perform training with 10, 100, 1,000 and 10,000 training levels on the three variants of mazes with increasing difficulties. The episode rewards on both the training and test sets are continuously observed along the training process. Showing in Figure 2a are the top 5 performing[1] agents with semi-transparent lines, and the best performing agents with bold solid lines. The associated test curves of those agents are plotted with the corresponding colors on the right side.

First of all, overfitting by "over-training" could be observed in a few agents with small training set. Although for many agents, especially the ones with larger training set, the test performances remain stable with very large numbers of training steps. Unsurprisingly, we observe that the *test* episode rewards increase with more training levels and decrease on more difficult mazes. On the other hand, the *training* episode rewards reach the optimal value in *all* cases. In other words, with the same (optimal) training rewards, the generalization performance could be very different depending on factors like the complexity of the mazes and training set size. In particular, the agents trained with only 10 training levels perform badly in all maze environments.

[1]Rankings are computed according to the average *training* rewards towards the end of the learning.

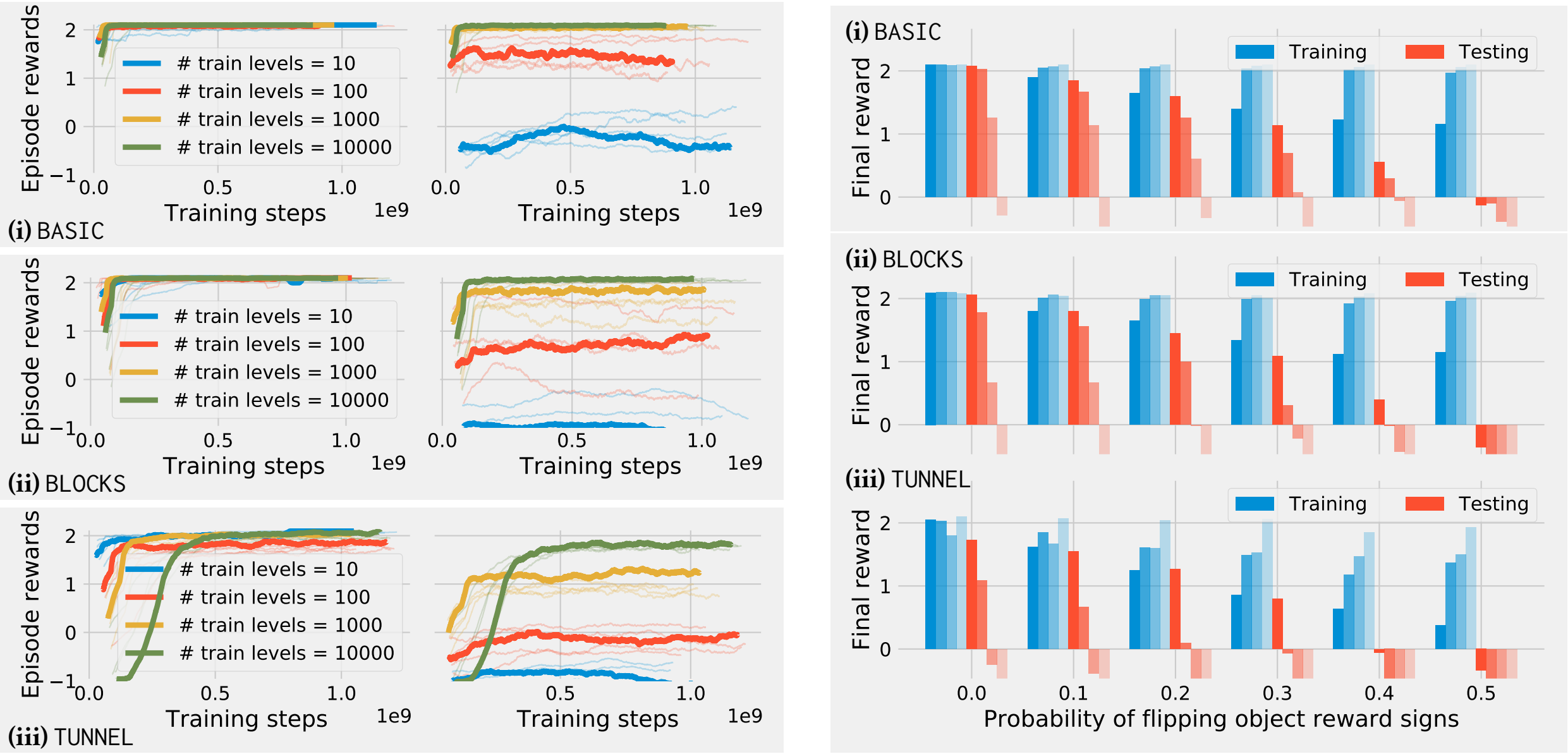

(a) Learning curves.

(b) Tests on randomized mazes.

Figure 2: (a) Learning curves. Training (left) and test (right) rewards on maze variants: (i) BASIC (ii) BLOCKS (iii) TUNNEL. (b) Experiments on (i) BASIC (ii) BLOCKS (iii) TUNNEL with random rewards. The horizontal axis indicates the probability that the sign of the reward of each object is flipped. The blue and red bars indicate the performance on the training and test levels. In each group, the four bars (from dark to light) show the results with 10,000, 1,000, 100, and 10 training levels, respectively.

### 3.3 Memorization Capacity of Deep RL Agents

Comparing to supervised learning, we observe that the neural networks used in deep RL are usually smaller, while the "effective number of samples", i.e. the number of different observed "states" (as oppose to "levels") by the agents, could be much larger. Could it be that the number of effective training samples is large enough comparing to the model complexity so that overfitting is less of an issue than in (deep) supervised learning? In this section, we explicitly evaluate the model capacity of our agents via a set of randomized games, following similar protocol as proposed for supervised learning in Zhang et al. (2017).

Specifically, we extend the game environment so that in the training levels, with probability $p$, the reward value $r$ of each object is flipped into $-r$ independently. The (pseudo) randomness is determined by the level id, so when the same level is encountered twice, the reward assignments are the same[2]. When $p = 0.5$, the sign of the reward of each object is positive or negative with equal probability. In this case, good *test* performance is impossible (by design), and good *training* performance means serious overfitting.

Figure 2b shows the results on different maze variants and with different $p$ (noise levels). Within each group of bars, the color intensity indicate the training set size. The lightest bars correspond to the case with 10 training levels and the darkest ones correspond to 10,000 training levels. For small training set sizes, the agents could reach near optimal training performance in all cases. The capacities of the networks start to be insufficient for memorizing all random reward assignments on larger training set, and the situation is more pronounced on the more difficult TUNNEL mazes. However, even for TUNNEL mazes with 10,000 training levels, the agents achieve non-trivial training rewards[3] under heavy noises, creating a big gap between the training and testing performances.

[2] In order for the episode rewards to be easily compared, we post-process the reward values by re-scaling so that the maximum achievable reward for each level is still 2.1. We discard levels with all-negative rewards. Besides, the random flipping are only applied to the training levels, the test levels are untouched.

In summary, the deep RL agents demonstrate the capability to memorize a non-trivial number of training levels even with completely random rewards. Therefore, the test performances for the same network architectures, learning methods and the same (near optimal) training rewards could be drastically different depending on the tasks.

## 4 Evaluation of Generalization in Deep RL

In the previous sections, we saw that depending on the training set size, the regularity of the tasks and many other factors, the performance on unseen test scenarios could be drastically different from the training. Therefore, the rewards achieved on the training set alone might not be very informative for out-of-sample test performances.

On the other hand, since the formal setup of RL problems does not typically have separate training and testing stages, it is common to see evaluation performances reported directly on the training environments. In order to avoid potentially misleading performances from seriously overfitted agents, various techniques are introduced to inject extra stochasticity to the environments during the evaluation process. See Table 1 for a few commonly used stochasticity-injection techniques in the Atari game evaluations. Adding stochasticity is popular because a simple way to achieve superficially high scores in a deterministic environment is to optimize for open-loop action sequences. With enough computation, algorithms like *trajectory tree* (Kearns et al., 2000) or *Brute* (Machado et al., 2017) could reach optimal rewards without even looking at the intermediate states. In fact, *Brute* outperformed the best learning method on 45 out of 55 Atari 2600 games at the time of the reports in Bellemare et al. (2015). But those methods are vulnerable to small perturbations and do not generalize to unseen data.

However, the wide variety of evaluation techniques leads to potential difficulties in directly comparing the performances from different papers (Machado et al., 2017) or reproducing specific results (Henderson et al., 2017). Moreover, the efficacy of different techniques depends heavily on the nature of specific tasks. In this section, we comparatively evaluate some widely used techniques in our framework. In particular, each technique can be used in two different ways: 1) as a regularizer to *prevent* overfitting; 2) as an evaluation add-on to *detect* overfitting. We present the results of 1) along with the definitions of the techniques studied here. The results for 2) follow. See Appendix B for more results.

**Stochastic policy** A policy is a function that maps the input state to an action. A stochastic policy does not choose action deterministically. It can be used to encourage exploration during training by allowing the agents to occasionally execute low-confidence actions. Stochastic policy is always assumed in the formulation of policy gradient based training algorithms. In our experiments, the policy network produces a multinomial distribution over the action space, from which the agent samples actions. Typically, an entropy regularizer is also used to discourage the networks from learning policies that are highly concentrated on a single action. However, from the results in Subsection 3.2, we can see that this amount of stochasticity (as regularization) does not prevent the agents from overfitting to random mazes.

**Null-op starts and human starts** A technique widely used in Atari evaluation is to put the agent in a different initial state during testing. This is typically achieved via starting by executing a number of no-ops (*null-op starts*), or taking over from the middle of a human gameplay trajectory (*human starts*) (Nair et al., 2015). Though it is sometimes criticized that the human gameplay could be highly biased and not necessarily representative for fair evaluation (Machado et al., 2017), due to limited control over the simulators, it is not easy to simply sample a random valid initial state in most of the Atari games. In our maze game, we can algorithmically generate random starting states by spawning the agent at a random initial location. In this setting, for the same level id, the wall configurations, locations and rewards of objects will be consistent, but the initial starting point of the agent could be different every time. We call this mode RAND-SPAWN. We apply it as a regularizer during training, and Figure 3a-(i) shows the performance under various noise levels and training set sizes on BASIC mazes. Comparing with Figure 2b-(i), we can see that random starts help to improve the test performances a bit. However, the agents are still able to fit even random training levels almost optimally.

[3] An untrained random agent is expected to get negative episode rewards due to the timeout penalty of $-1$.

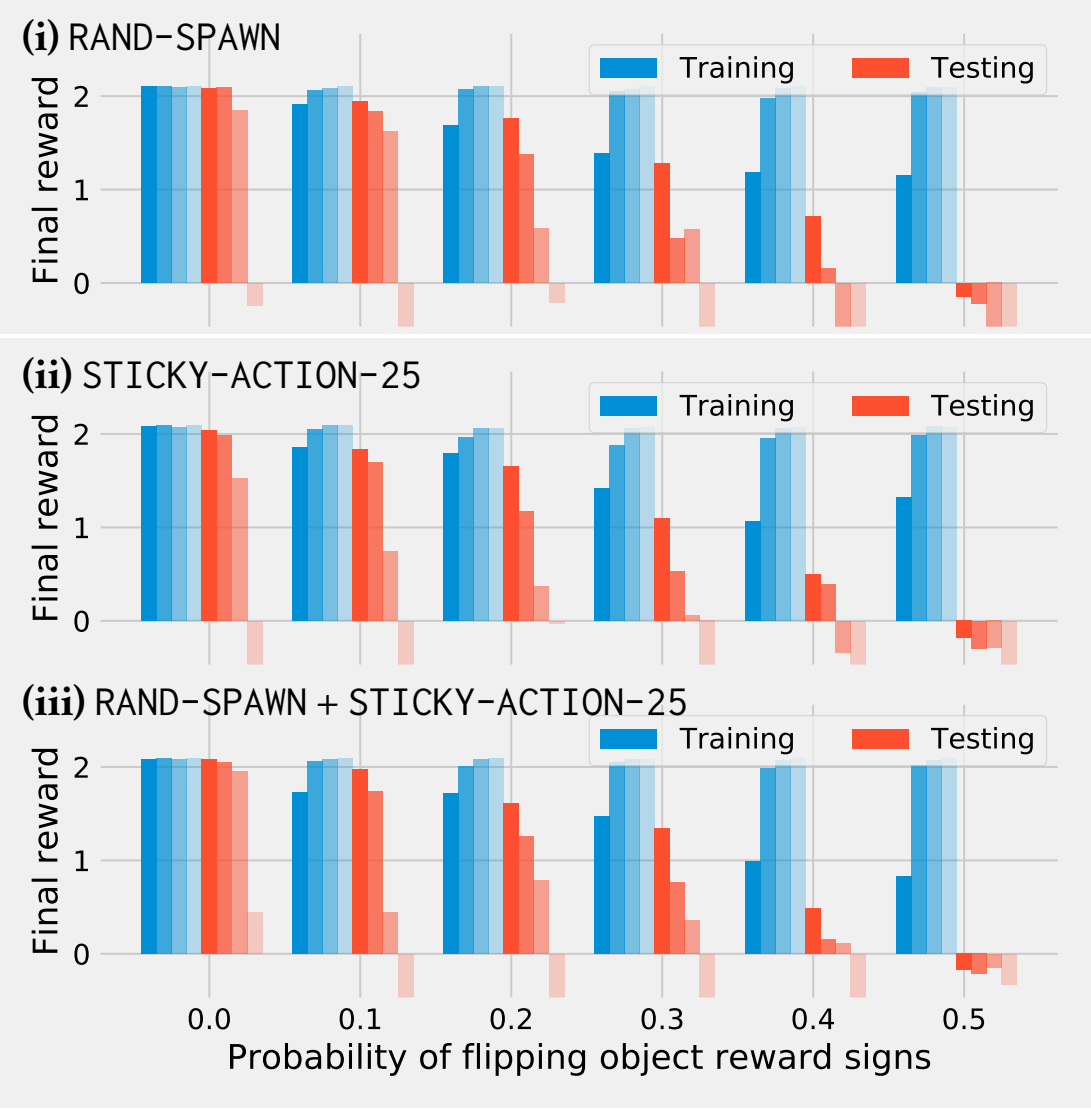

(a) Stochasticity as regularizers.

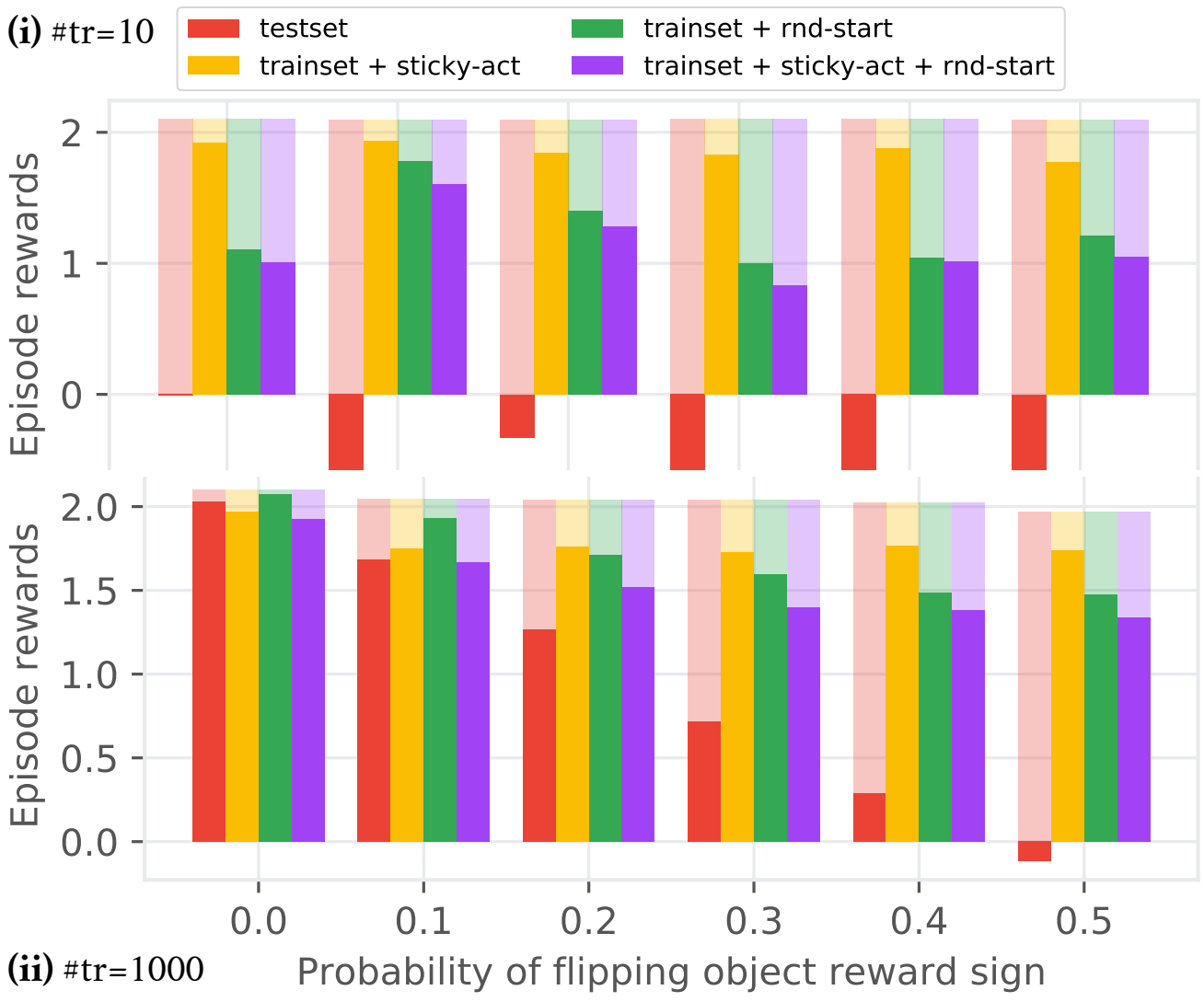

(b) Stochasticity as evaluation.

Figure 3: (a) Evaluation of the effectiveness of regularizers (i) random starts (ii) sticky action ($\zeta = 0.25$) (iii) both combined, on BASIC mazes. The bar plots are arranged the same way as in Figure 2b. (b) Evaluating with different protocols on BASIC mazes, with (i) 10 (ii) $10^3$ training levels. The transparent bars show the training rewards. Training is *without* random starts or sticky actions.

**Sticky actions** With a *stickiness* parameter $\zeta$, at time $t$, the action $A_t$ that the environment executes is

$$A_t = \begin{cases} a_t, & \text{with prob. } 1 - \zeta \\ a_{t-1}, & \text{with prob. } \zeta \end{cases}$$

where $a_t$ is the action proposed by the agent at time $t$ (see Appendix D for an alternative definition). It is proposed as the preferred way to add stochasticity in Machado et al. (2017) and shown to effectively detect overfittings of *Brute*. We apply sticky actions as a regularizer during training, with the stickiness $\zeta = 0.25$ following Machado et al. (2017), and report the results on BASIC in Figure 3a-(ii), and also in Figure 3a-(iii) the results combining sticky actions and random starts. The results look very similar to random starts alone.

When used as evaluation criterions, we train the agents *without* those regularizers, but apply them to add stochasticity when evaluating on the *training set*. The results are shown in Figure 3b. The evaluations on the *test set* are also shown as reference. As we can see, the agents perform poorly on the test set when the training set is small or when the game is random, yet for those overfitted agents, the evaluation scores on the training set with sticky actions, random starts or both remain high. Figure 4 illustrates how sticky actions make only small perturbations to the states, and ConvNets based agents are automatically robust to it. It also shows how sticky actions with different levels of stickiness all fail to distinguish agents with drastic generalization performances on the test set. Fore more details, see Appendix C for detailed analysis of trajectories and how the agents are robust to other stochasticity even without explicit regularization, and Appendix D for results on different stickiness modes.

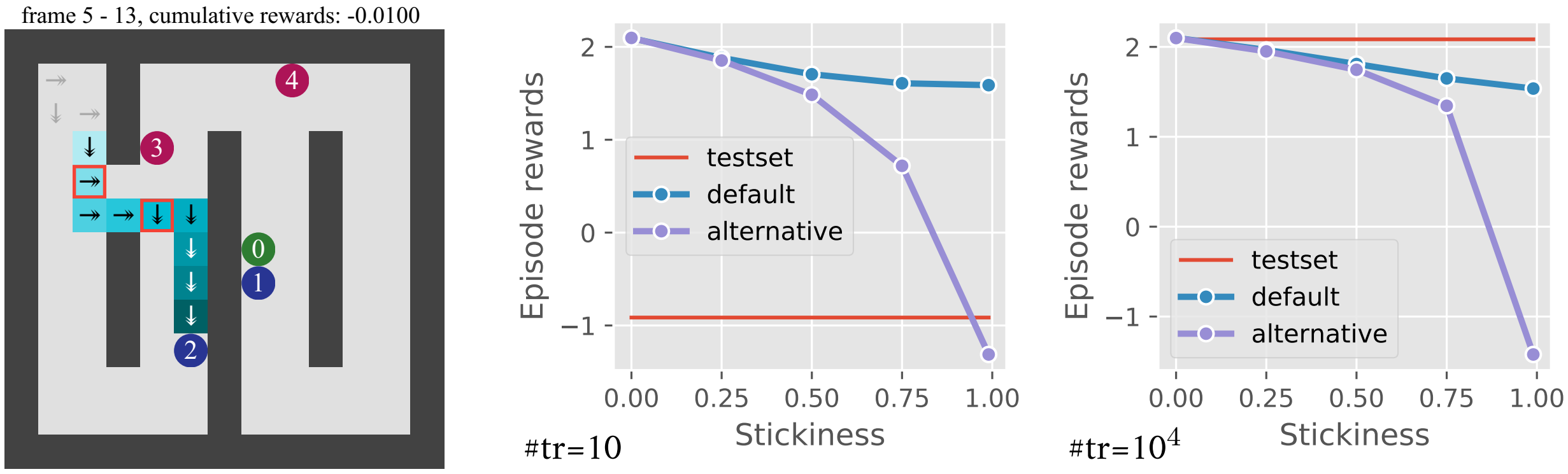

Figure 4: Left: example of a segment of trajectory with sticky actions: red frames indicate that the proposed action is different from the executed action. Right: Evaluation performances based on different stickiness levels and types (default vs alternative) of sticky actions. Red lines show the performance on the test set: the case with 10 training levels performs much worse than with $10^4$ training levels, but evaluations with sticky actions barely show the difference. Please see Appendix C and D for more details.

In summary, while stochasticity is effective on algorithms like *Brute* that exploit determinism, it could neither *prevent* deep RL agents from serious overfitting nor *detect* overfitted agents effectively when evaluated on the training set.

## 5 Discussions on Overfitting in RL

With complicated learning dynamics, overfitting in RL could happen in many different forms and should be treated very carefully. Algorithms that exploit determinism of the environment by ignoring the states and memorizing effective open-loop action sequences could attain state-of-the-art scores (Braylan et al., 2015; Bellemare et al., 2015). Those agents are generally considered overfitted as they are sensitive to small perturbations in the environments and could not generalize to unseen states. However, as shown in Section 4, memorization could still happen in stochastic environments. In particular, an agent does not need to predict every action "correctly", as long as it achieves the optimal rewards in the end. As is shown, an agent could fit to random noises robustly in the task even when trained with regularizers that explicitly add stochasticity to the environment.

Moreover, as shown in Figure 3b, an agent might overfit robustly even without explicit regularization, resulting in artificially high rewards when evaluated on the training set with added stochasticity. Random starts and sticky actions, while effective at identifying determinism-exploiting algorithms like *Brute* (Machado et al., 2017), might not be very effective at *detecting* overfitting for agents with neural network function approximators trained via gradient methods. The behaviors of those blackbox policies are relatively poorly understood, and they might implicitly acquire certain kind of robustness due to the architectures or the training dynamics. See Appendix C for detailed trajectory-level analysis of potential reasons for the implicit robustness.

A closely related notion that is widely studied in RL is the *exploration-exploitation trade-off* (Sutton & Barto, 1998). The issue of balancing between exploration and exploitation does not arise in supervised learning as the inputs are independent and identically distributed (i.i.d.) samples and the learner do not affect how they are generated. It also does not directly map to either overfitting or underfitting. An overly exploited agent could get highly specialized in a subset of the state space so that it performs badly when encountering unfamiliar states. This resembles the behavior of overfitting. But insufficient exploration could also trap an agent in a locally optimal strategy without knowing alternative policies that achieve higher training rewards. Since the training performance is sub-optimal, it behaves like underfitting, or optimization failure in supervised learning.

In fact, gradient based optimization algorithms without convergence guarantees are extensively used in deep RL. Even with sufficient exploration, an agent could still learn sub-optimal models due to inferior optimization. Despite many successful applications, optimization in deep RL tends to be more difficult than in deep supervised learning. For example, Mnih et al. (2013) report that the average episodic reward curves against the training steps are very unstable. Henderson et al. (2017) point out that even different random seeds for neural network weight initialization could affect the learning results drastically.

On the other hand, memorization and overfitting might not always be undesirable. In our everyday life, humans constantly "overfit" some subroutines via "muscle memories". Those subroutines allow us to act efficiently without needing to consult the brain with every single detail, but they usually do not generalize well for even small perturbations. A key ability of us is to detect the failures of an overfitted component and invoke higher level intelligence to adapt or re-learn it. The efforts needed for re-learning depend on the complexity of the subroutine. An interesting example is the so called "backwards brain bicycle", which is a regular bike that reverses how the handlebars operates. The front wheel will turn right if you turn the handlebar left, and vice versa[4]. This single modification could cost us up to several months (Sandlin, 2015) to re-learn the bike-riding subroutine.

In summary, in complicated learning scenarios, overfitting could happen in various forms. Studying overfitting not only allows us to better understand the behaviors of the learning algorithms, but can also be potentially useful to foster hierarchical learning systems. Recent advances in theoretical analysis of the effective complexity of trained neural networks have improved our understandings of generalization in supervised learning. In the next section, we provide a study in deep RL from the perspective of inductive bias.

Practically, to detect overfittings, it is recommended to use separate training and validating / testing sets that are statistically tied. Developments of standardized protocols, benchmarks and evaluation platforms for common categories of RL problems that could effectively assess the generalization performances and identify potential overfittings will be vital for the advances of the field. An example of exciting progresses in this direction is the OpenAI *Gym Retro*, which is a new RL platform for classic games like SEGA Genesis games, announced during the writing of this paper. It supports proper training / test split with evaluation scores reported on separate test sets.

## 6 Inductive Bias of Algorithms and Problems

The experiments in this paper show that our deep RL agents have large capacity to memorize random games in the training set; on the other hand, with the setup, they could generalize well on regular mazes. Similar phenomenons are reported in supervised learning (Zhang et al., 2017), and recent work shows various approaches to explain this (Bartlett et al., 2017; Neyshabur et al., 2017; Advani & Saxe, 2017; Smith & Le, 2017; Dziugaite & Roy, 2017; Liang et al., 2017). The main intuition is that, even with the same hypothesis space and training algorithm, the trained networks could be quite different depending on the data distributions of the underlying tasks. Therefore, complexity measures for the trained models (instead of the whole *hypothesis space*) should be measured for tighter generalization bounds. However, it remains largely open to characterize what properties of tasks could lead to simple trained models in deep learning. In deep RL, the learning dynamics is even more complicated than supervised learning because the samples are not i.i.d.

Inductive bias is a notion for the *a priori* algorithmic preferences. For neural networks, the architectures typically account for a big part of the inductive bias of the learning process. Carefully designed architectures could significantly outperform vanilla densely connected networks in specific tasks. For example, with spatial locality built in the structures, ConvNets are shown to be very effective in difficult computer vision problems.

[4]This can be "simulated" on a regular bike by riding cross-handed. Please do not try at home without proper safety measures.

Table 2: Inductive bias of MLPs and ConvNets. Each pair of numbers $r_{\text{MLP}}/r_{\text{ConvNet}}$ in the table shows the final episode rewards achieved by the average of top-3 MLP based agents and the average top-3 ConvNet based agents, respectively. The colored dots help visualize the two numbers, with red being large and blue being small. Each column corresponds to a setup with different probability of flipping the rewards. The numbers in the "Testing" block shows the performance on the held-out test levels of the corresponding agents in the "Training" block. Each row indicates the setup with different number of training levels.

| # train levels \ flip p | Training | | | | Testing | | | |
|---|---|---|---|---|---|---|---|---|
| | 0.0 | 0.2 | 0.4 | 0.5 | 0.0 | 0.2 | 0.4 | 0.5 |
| $10^1$ | 2.1/ 2.1 | 2.1/ 2.1 | 2.1/ 2.1 | 2.1/ 2.1 | -0.6/ -0.1 | -0.2/ -0.3 | -0.4/ -0.8 | -0.4/ -0.6 |
| $10^2$ | 2.1/ 2.1 | 2.1/ 2.1 | 2.1/ 2.1 | 2.0/ 2.1 | 0.9/ 1.2 | 0.4/ 0.9 | 0.0/ -0.3 | -0.4/ -0.6 |
| $10^3$ | 2.1/ 2.1 | 2.0/ 2.0 | 2.0/ 2.0 | 2.0/ 2.0 | 1.6/ 2.0 | 0.9/ 1.1 | 0.3/ 0.3 | -0.2/ -0.1 |
| $10^4$ | 2.1/ 2.1 | 1.7/ 1.6 | 1.3/ 1.2 | 1.3/ 1.2 | 1.9/ 2.1 | 1.4/ 1.5 | 0.8/ 0.5 | -0.1/ -0.1 |

Table 3: Performance comparison between agents using MLPs and Big-ConvNets. Each pair of numbers $r_{\text{MLP}}/r_{\text{Big-ConvNet}}$ shows the final episode rewards for the average top-3 MLP-based agents and the average top-3 Big-ConvNet-based agents, respectively. Big-ConvNets have more parameters than normal ConvNets used in the rest of the paper (see Appendix A.3). Formatted similarly as Table 2.

| # train levels \ flip p | Training | | | | Testing | | | |
|---|---|---|---|---|---|---|---|---|
| | 0.0 | 0.2 | 0.4 | 0.5 | 0.0 | 0.2 | 0.4 | 0.5 |
| $10^1$ | 2.1/ 2.1 | 2.1/ 2.1 | 2.1/ 2.1 | 2.1/ 2.1 | -0.6/ -0.3 | -0.2/ -0.3 | -0.4/ -0.2 | -0.4/ -0.9 |
| $10^2$ | 2.1/ 2.1 | 2.1/ 2.0 | 2.1/ 2.1 | 2.0/ 2.1 | 0.9/ 1.4 | 0.4/ 1.1 | 0.0/ 0.1 | -0.4/ -0.4 |
| $10^3$ | 2.1/ 2.1 | 2.0/ 2.0 | 2.0/ 2.0 | 2.0/ 2.0 | 1.6/ 2.0 | 0.9/ 1.2 | 0.3/ 0.5 | -0.2/ -0.0 |
| $10^4$ | 2.1/ 2.1 | 1.7/ 1.8 | 1.3/ 1.6 | 1.3/ 1.6 | 1.9/ 2.1 | 1.4/ 1.7 | 0.8/ 0.6 | -0.1/ 0.1 |

For sequence data, Long Short Term Memory (LSTMs) and *attention* mechanisms are commonly used. More sophisticated network structures like *differentiable memory* can also be found in RL tasks.

To achieve good generalization, it is important that the inductive bias of the algorithms is compatible with the bias of the problems. In particular, comparing the default setting of the maze with the one with randomized rewards, a clear regularity property of the formal is that the rewards are *spatially invariant*. In other words, the reward value of an object is constant, regardless of where it is placed by the maze level generator. On the other hand, as random flipping is introduced, the reward value is no longer invariant to its location. Instead, it will be tied with the initial configuration of the maze, which is determined by the random seed associated with a particular level. To compare how the task bias interacts with the model inductive bias, we train two different sets of deep RL agents: one based on convolutional neural networks (ConvNets), while the other based on multi-layer perceptrons (MLPs). ConvNets encode the bias of spatial invariance by re-using local filters across the spatial domain. On the other hand, MLPs use dense connections between layers of neurons. Apart from the same ConvNets as used in other experiments of this paper, we also test on bigger ConvNets with much larger capacities. See Appendix A.3 for the details of the network architectures tested.

The results on `BASIC` mazes with varying training set sizes and reward-flipping probabilities are shown in Table 2. Each pair of numbers in the table shows the final episode rewards of MLP-based agents and ConvNet-based agents, respectively. Our main observations are: 1) When the tasks are highly regular, both MLPs and ConvNets achieve optimal *training* rewards; 2) The *training* performance of MLPs tend to outperform ConvNets, especially on more randomized tasks and large training set sizes; 3) Looking at the *test* performance, we can see that ConvNets consistently outperform MLPs when the games are regular, and as expected, both perform badly on random games.

In summary, the MLPs are better at fitting the training levels, but generalize worse than the ConvNets. When the rewards are spatially invariant, the ConvNets generalize much better than MLPs. On the other hand, ConvNets, given enough weights, are also (approximately) universal. They are shown to be able to recover the original inputs from hidden representations (Bruna et al., 2014; Gilbert et al., 2017) and memorize random labels for images (Zhang et al., 2017). In Table 3, we compare MLPs and ConvNets with higher capacity (Big-ConvNets) under the same settings. As we can see, with added capacity, the ConvNets are no longer underperforming the MLPs on the *training* levels even with the heaviest randomization. On the other hand, it still generalize significantly better than MLPs on low noise regular games, potentially due to the compatible inductive bias.

In conclusion, even though Big-ConvNets have the same or higher capacity than MLPs at memorizing random mazes, they generalize better on regular games with spatial invariant rewards. The empirical observations suggest that a notion of compatibility of the inductive bias of the algorithms / models with the bias of the problems could be a good direction to study the generalization behaviors. However, defining formal characterizations that are both mathematically easy to manipulate and practically encompass a wide range of real world problems is still an open problem.

# 7 Conclusion

Large neural networks, together with powerful training algorithms are highly effective at memorizing a large (random) training set similar to the case of supervised learning. When the algorithmic inductive bias matches the task well, good generalization performance could still be obtained. However, formal characterization of the inductive bias is challenging, and theoretical understanding of the generalization performance of trained over-parameterized agents is still largely open.

In practice, evaluation protocols need to be carefully designed to be able to detect overfiting. The effectiveness of stochasticity-based evaluation techniques highly depends on the properties of the tasks, as the agents could still implicitly learn to overfit robustly. Therefore, an isolation of the training and test data is recommended even for noisy and non-deterministic environments.

**Acknowledgments** The authors would like to thank Neil Rabinowitz, Eric Jang and David Silver for helpful discussions and comments.

# References

Advani, M. S and Saxe, A. M. High-dimensional dynamics of generalization error in neural networks. *arXiv:1710.03667*, 2017.

Agarwal, A., Hsu, D., Kale, S., Langford, J., Li, L., and Schapire, R. Taming the monster: A fast and simple algorithm for contextual bandits. In *ICML*, 2014.

Auer, P. Using upper confidence bounds for online learning. In *Annual Symposium on Foundations of Computer Science*, pp. 270–279, 2000.

Auer, P., Cesa-Bianchi, N., and Fischer, P. Finite-time analysis of the multiarmed bandit problem. *Machine Learning*, 47(2):235–256, May 2002. ISSN 1573-0565.

Azar, M. G., Osband, I., and Munos, R. Minimax regret bounds for reinforcement learning. In *ICML*, volume 70, pp. 263–272, 06–11 Aug 2017.

Bartlett, P. L, Foster, D. J, and Telgarsky, M. J. Spectrally-normalized margin bounds for neural networks. In *NIPS*, pp. 6241–6250, 2017.

Bellemare, M. G., Naddaf, Y., Veness, J., and Bowling, M. The arcade learning environment: An evaluation platform for general agents (extended abstract). In *IJCAI*, 2015.

Braylan, A., Hollenbeck, M., Meyerson, E., and Miikkulainen, R. Frame skip is a powerful parameter for learning to play atari. In *AAAI-15 Workshop on Learning for General Competency in Video Games*, 2015.

Brockman, G., Cheung, V., Pettersson, L., Schneider, J., Schulman, J., Tang, J., and Zaremba, W. OpenAI Gym, 2016.

Bruna, J., Szlam, A., and LeCun, Y. Signal recovery from pooling representations. In *ICML*, 2014.

Bubeck, S. and Cesa-Bianchi, N. Regret analysis of stochastic and nonstochastic multi-armed bandit problems. *Foundations and Trends in Machine Learning*, 5(1):1–122, 2012.

Chakraborty, B. and Murphy, S. A. Dynamic treatment regimes. *Annual review of statistics and its application*, 1: 447–464, 2014.

Dann, C., Lattimore, T., and Brunskill, E. Unifying PAC and regret: Uniform PAC bounds for episodic reinforcement learning. In *NIPS*, pp. 5717–5727, 2017.

Deng, Y., Bao, F., Kong, Y., Ren, Z., and Dai, Q. Deep direct reinforcement learning for financial signal representation and trading. *IEEE transactions on neural networks and learning systems*, 28(3):653–664, 2017.

Dziugaite, G. K. and Roy, D. M. Computing nonvacuous generalization bounds for deep (stochastic) neural networks with many more parameters than training data. In *UAI*, 2017.

Gilbert, A. C, Zhang, Y., Lee, K., Zhang, Y., and Lee, H. Towards understanding the invertibility of convolutional neural networks. In *IJCAI*, 2017.

Glavic, M., Fonteneau, R., and Ernst, D. Reinforcement learning for electric power system decision and control: Past considerations and perspectives. In *World Congress of the International Federation of Automatic Control, Toulouse 9-14 July 2017*, pp. 1–10, 2017.

Hardt, M., Recht, B., and Singer, Y. Train faster, generalize better: Stability of stochastic gradient descent. In *ICML*, volume 48, pp. 1225–1234, 20–22 Jun 2016.

Hausknecht, M. and Stone, P. The impact of determinism on learning atari 2600 games. In *AAAI Workshop on Learning for General Competency in Video Games*, 2015.

He, K., Zhang, X., Ren, S., and Sun, J. Deep residual learning for image recognition. In *CVPR*, pp. 770–778, 2016.

Henderson, P., Islam, R., Bachman, P., Pineau, J., Precup, D., and Meger, D. Deep reinforcement learning that matters. *arXiv:1709.06560*, 2017.

Huang, S., Papernot, N., Goodfellow, I., Duan, Y., and Abbeel, P. Adversarial attacks on neural network policies. *arXiv:1702.02284*, 2017.

Jaksch, T., Ortner, R., and Auer, P. Near-optimal regret bounds for reinforcement learning. *JMLR*, 11(Apr): 1563–1600, 2010.

Jiang, N., Krishnamurthy, A., Agarwal, A., Langford, J., and Schapire, R. E. Contextual decision processes with low Bellman rank are PAC-learnable. In *ICML*, 2017.

Kearns, M. J, Mansour, Y., and Ng, A. Y. Approximate planning in large pomdps via reusable trajectories. In *NIPS*, pp. 1001–1007, 2000.

Kosorok, M. R and Moodie, E. EM. *Adaptive treatment strategies in practice: planning trials and analyzing data for personalized medicine*. SIAM, 2015.

Lattimore, T. and Hutter, M. Near-optimal PAC bounds for discounted MDPs. *TCS*, 558:125–143, 2014.

Li, Y., Szepesvari, C., and Schuurmans, D. Learning exercise policies for american options. In *Artificial Intelligence and Statistics*, pp. 352–359, 2009.

Liang, T., Poggio, T., Rakhlin, A., and Stokes, J. Fisher-rao metric, geometry, and complexity of neural networks. *arXiv:1711.01530*, 2017.

Lichman, M. UCI machine learning repository, 2013. URL `http://archive.ics.uci.edu/ml`.

Lillicrap, T. P, Hunt, J. J, Pritzel, A., Heess, N., Erez, T., Tassa, Y., Silver, D., and Wierstra, D. Continuous control with deep reinforcement learning. In *ICLR*, 2016.

Lin, J., Camoriano, R., and Rosasco, L. Generalization properties and implicit regularization for multiple passes SGM. *ICML*, pp. 2340–2348, 2016.

Machado, M. C, Bellemare, M. G, Talvitie, E., Veness, J., Hausknecht, M., and Bowling, M. Revisiting the arcade learning environment: Evaluation protocols and open problems for general agents. *arXiv:1709.06009*, 2017.

Mannion, P., Duggan, J., and Howley, E. An experimental review of reinforcement learning algorithms for adaptive traffic signal control. In *Autonomic Road Transport Support Systems*, pp. 47–66. Springer, 2016.

Mnih, V., Kavukcuoglu, K., Silver, D., Graves, A., Antonoglou, I., Wierstra, D., and Riedmiller, M. Playing atari with deep reinforcement learning. *arXiv:1312.5602*, 2013.

Mnih, V., Kavukcuoglu, K., Silver, D., Rusu, A. A, Veness, J., Bellemare, M. G, Graves, A., Riedmiller, M., Fidjeland, A. K, Ostrovski, G., et al. Human-level control through deep reinforcement learning. *Nature*, 518(7540): 529–533, 2015.

Mnih, V., Badia, A. P., Mirza, M., Graves, A., Lillicrap, T., Harley, T., Silver, D., and Kavukcuoglu, K. Asynchronous methods for deep reinforcement learning. In *ICML*, pp. 1928–1937, 2016.

Nair, A., Srinivasan, P., Blackwell, S., Alcicek, C., Fearon, R., De Maria, A., Panneershelvam, V., Suleyman, M., Beattie, C., Petersen, S., et al. Massively parallel methods for deep reinforcement learning. *arXiv:1507.04296*, 2015.

Neyshabur, B., Bhojanapalli, S., McAllester, D., and Srebro, N. Exploring generalization in deep learning. In *NIPS*, pp. 5949–5958, 2017.

Nouri, A., Littman, M. L, Li, L., Parr, R., Painter-Wakefield, C., and Taylor, G. A novel benchmark methodology and data repository for real-life reinforcement learning. In *Multidisciplinary Symposium on Reinforcement Learning*, 2009.

Raghu, M., Irpan, A., Andreas, J., Kleinberg, R., Le, Q. V, and Kleinberg, J. Can deep reinforcement learning solve erdos-selfridge-spencer games? *arXiv:1711.02301*, 2017.

Sandlin, D. The backwards brain bicycle, 2015. URL `https://www.youtube.com/watch?v=MFzDaBzBlL0`.

Silver, D., Huang, A., Maddison, C. J, Guez, A., Sifre, L, Van Den Driessche, G., Schrittwieser, J., Antonoglou, I., Panneershelvam, V., , et al. Mastering the game of go with deep neural networks and tree search. *Nature*, 529 (7587):484–489, 2016.

Silver, D., Hubert, T., Schrittwieser, J., Antonoglou, I., Lai, M., Guez, A., Lanctot, M., Sifre, L., Kumaran, D., Graepel, T., et al. Mastering chess and shogi by self-play with a general reinforcement learning algorithm. *arXiv:1712.01815*, 2017a.

Silver, D., Schrittwieser, J., Simonyan, K., Antonoglou, I., Huang, A., Guez, A., Hubert, T., Baker, L., Lai, M., Bolton, A., et al. Mastering the game of go without human knowledge. *Nature*, 550(7676):354, 2017b.

Smith, S. and Le, Q. Understanding generalization and stochastic gradient descent. *arXiv:1710.06451*, 2017.

Strehl, A. L, Li, L., and Littman, M. L. Reinforcement learning in finite MDPs: PAC analysis. *JMLR*, 10(Nov): 2413–2444, 2009.

Sutton, R. S and Barto, A. G. *Reinforcement learning: An introduction*. MIT press Cambridge, 1998.

van der Pol, E. and Oliehoek, F. A. Coordinated Deep Reinforcement Learners for Traffic Light Control. In *NIPS*, 2016.

Vinyals, O., Ewalds, T., Bartunov, S., Georgiev, P., Vezhnevets, A., Yeo, M., Makhzani, A., Küttler, H., Agapiou, J., Schrittwieser, J., et al. StarCraft II: a new challenge for reinforcement learning. *arXiv:1708.04782*, 2017.

Wen, Z., O'Neill, D., and Maei, H. Optimal demand response using device-based reinforcement learning. *IEEE Transactions on Smart Grid*, 6(5):2312–2324, 2015.

Whiteson, S., Tanner, B., Taylor, M. E, and Stone, P. Protecting against evaluation overfitting in empirical reinforcement learning. In *Adaptive Dynamic Programming And Reinforcement Learning (ADPRL)*, pp. 120–127. IEEE, 2011.

Yu, Y.-L., Li, Y., Schuurmans, D., and Szepesvári, C. A general projection property for distribution families. In *NIPS*, pp. 2232–2240, 2009.

Zhang, C., Bengio, S., Hardt, M., Recht, B., and Vinyals, O. Understanding deep learning requires rethinking generalization. In *ICLR*, 2017.

# A Experiment Details

## A.1 Task Setup

The maze game in this paper is a 2D gridworld environment. On the map, there are 5 different objects. The agent's goal is to collect the objects with positive rewards while avoiding the objects with negative rewards, and then collect the *terminating object* to end the episode before the forced termination with a timeout penalty is triggered. The BASIC version of the map is a $9 \times 9$ empty room; the BLOCKS version is a $13 \times 13$ room with 8 randomly placed single-block obstacles; the TUNNEL version is also $13 \times 13$, with randomly generated continuous internal walls that form tunnel-like maps. Figure 1a shows an example of the generated TUNNEL maze, with objects and agent placed. The potentially long corridors make it more difficult to navigate than the other two variants. The map generator ensures that the maze is solvable — there is at least one valid route to collect all objects with positive rewards and the terminating object.

In the default setup, taking object 1 (*red diamond*) and 2 (*blue diamond*) will give the agent $+1$ reward each, while object 3 (*bomb*) and 4 (*thunder*) will give $-1$ reward each. Object 0 (*key*) is a *terminating object* with a reward 0.1, which ends the episode immediately when taken. If the agent fails to take the terminating object within 200 steps, the episode ends with a timeout penalty of $-1$. No moving cost is enforced, but the agent receives a penalty of $-0.01$ if attempting to step into an obstacle or the wall. As a result, an oracle agent could achieve an episode reward of 2.1. On the other hand, an un-trained random agent is expected to get negative episode rewards due to the timeout penalty.

In the randomization tests studied in Subsection 3.3, the mazes are modified by randomly flipping the sign of the reward of each object. Specifically, for a noise level $p$, independently for each object with reward $r$, we re-assign the reward to $-r$. To allow easy comparison of the results, we keep the maximum episode reward 2.1 by rescaling the reward values after the sign flips. If all the objects are with negative rewards, it is impossible to rescale, then we simple discard this configuration and re-sample random sign flippings. The sign flippings are kept consistent for a given level by using the level id as the random seed.

The input to the agent is a multi-channel "image" with the maze wall, location of each object and the location of the agent itself encoded in separate channels. Standard *Asynchronous Actor-Critic Agents* (A3C) is used in learning, with a policy head of 5 discrete actions including UP, DOWN, LEFT, RIGHT and STAY.

## A.2 Training Protocol and Dataset Splitting

We use the standard A3C training procedures, where each worker interacts with an independent copy of the game environment, and asynchronously update the central parameters. Because the games have finite number of steps, we follow an episodic setting. Each time an episode ends, the environment is reset and reinitialized with a new random maze configuration. The *episode reward* refers to the total rewards collected in a full episode.

To mimic the behavior of typical video games, which are commonly used to test RL algorithms, we also include the notion of *levels*. The level id acts as pseudo random number generator seed, which determines the initial state for the corresponding level. Furthermore, as video games normally contains a finite number of levels, we simulate the behavior by reserving a pool of finite number of ids for the training levels, and a separate pool for the testing levels. Note there is *no* explicit ranking of difficulties among different levels in our settings.

In RL, there is usually no explicit test stage formulated. Instead, the problem is formulated in a continuous learning setting where the goal is to maximize the cumulative and discounted rewards. But in a finite-horizon discrete world with finitely many (training) levels, the agent will eventually see repeated levels as training goes on, and potentially could start "overfitting" — attaining high episode rewards in the learning curves while performing poorly in levels that are not in the training set. Therefore, we reserve a test set of level ids that are disjoint from the training set, and during the A3C training, maintain a *test worker* that continuously evaluate the current agent on the test set. The test worker is almost identical to the other A2C workers in the A3C agent, except that it does *not* compute the gradients to update the parameters, and it runs on the pool of test level ids. Figure 1b illustrates the protocol. With this setup, we can not only evaluate the agent in separate clean environments, but also continuously observe its performance throughout the training process, generating similar learning curves that are commonly seen in supervised learning. However, as also noted in some previous work, the learning dynamics in RL are much less stable than in common supervised learning problems (Mnih et al., 2013; Henderson et al., 2017). Therefore, in our learning curve figures (e.g. Figure 2a), we use average smoothing with a window 200 to help visualization.

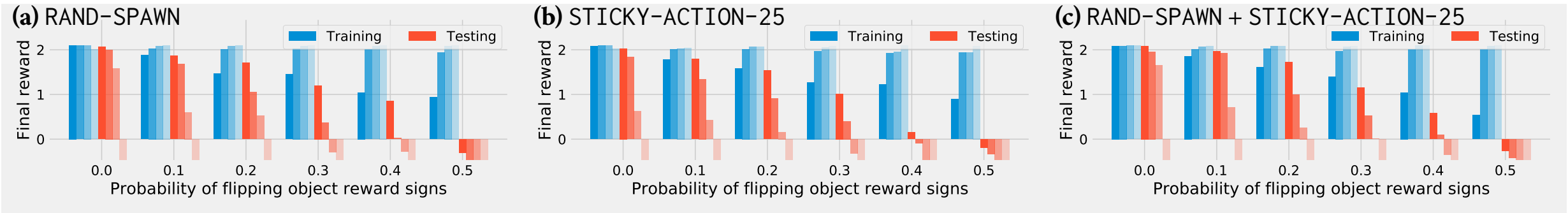

Figure 5: Evaluation of the effectiveness of regularizers (a) random starts (b) sticky action ($\zeta = 0.25$) (c) both combined, on BLOCKS mazes. The bar plots are arranged the same way as in Figure 2b.

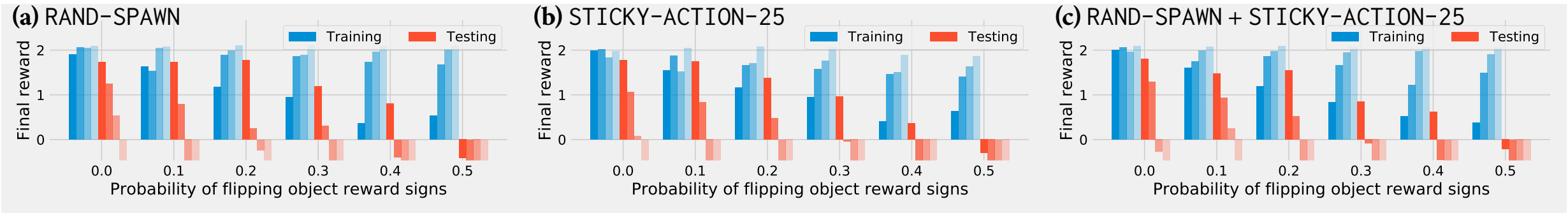

Figure 6: Evaluation of the effectiveness of regularizers (a) random starts (b) sticky action ($\zeta = 0.25$) (c) both combined, on TUNNEL mazes. The bar plots are arranged the same way as in Figure 2b.

### A.3 Network Architectures and Hyperparameters

Unless otherwise specified, the agents are using Convolutional Neural Networks (ConvNets) as function approximators. During the study of inductive bias in Section 6, we also tested Convolutional Networks with higher capacity (Big-ConvNets), and Multi-layer Perceptrons (MLPs). As noted in Henderson et al. (2017), learning in RL could sometimes fail completely due to unlucky random initializations. To filter out bad failure cases, for each specific setting, we run 20 different training jobs by sampling from a pool of pre-defined neural network architectures, and report top-k results as measured on the *training* levels. The specific network architectures are described below. The Rectified Linear Unit (ReLU) activation functions are used in all architectures.

**ConvNets** A random architecture from the followings:

- Conv($K_3 S_1 C_{11}$) $\rightarrow$ Conv($K_3 S_2 C_{11}$);
- Conv($K_3 S_1 C_{64}$) $\rightarrow$ Conv($K_3 S_2 C_{64}$);
- Conv($K_2 S_1 C_{64}$)$\times 2$ $\rightarrow$ Conv($K_3 S_2 C_{64}$);
- Conv($K_2 S_1 C_{64}$)$\times 3$ $\rightarrow$ Conv($K_3 S_2 C_{64}$);
- Conv($K_2 S_1 C_{64}$)$\times 4$ $\rightarrow$ Conv($K_3 S_2 C_{64}$).

The notation "Conv($K_2 S_1 C_{64}$))$\times 2$" means 2 consecutive convolutional layers with $2 \times 2$ kernels, stride 1 and 64 filter channels.

**Big-ConvNets** Similar to ConvNets, but with more layers and channels:

- Conv($K_3S_1C_{11}$) $\rightarrow$ Conv($K_3S_2C_{11}$);
- Conv($K_3S_1C_{64}$) $\rightarrow$ Conv($K_3S_2C_{64}$);
- Conv($K_2S_1C_{128}$)$\times 3$ $\rightarrow$ Conv($K_3S_2C_{128}$);
- Conv($K_2S_1C_{128}$)$\times 6$ $\rightarrow$ Conv($K_3S_2C_{128}$);
- Conv($K_2S_1C_{256}$)$\times 3$ $\rightarrow$ Conv($K_3S_2C_{256}$);
- Conv($K_2S_1C_{256}$)$\times 6$ $\rightarrow$ Conv($K_3S_2C_{256}$);
- Conv($K_2S_1C_{512}$)$\times 3$ $\rightarrow$ Conv($K_3S_2C_{512}$);
- Conv($K_2S_1C_{512}$)$\times 6$ $\rightarrow$ Conv($K_3S_2C_{512}$).

**MLPs** Densely connected neural networks with 2 to 3 hidden layers:

- Dense(512) $\rightarrow$ Dense(128);
- Dense(512) $\rightarrow$ Dense(512);
- Dense(1024) $\rightarrow$ Dense(1024);
- Dense(512) $\rightarrow$ Dense(128) $\rightarrow$ Dense(64);
- Dense(1024) $\rightarrow$ Dense(512) $\rightarrow$ Dense(128);
- Dense(1024) $\rightarrow$ Dense(1024) $\rightarrow$ Dense(1024);

For the A3C training, with the modified protocol illustrated in Figure 1b, we use 4 training workers and 1 test worker that does not update the parameters. We clip the reward values to be within $[-2, 2]$, and use an unroll length of 15 steps. An entropy regularizer with coefficient log-uniformly sampled from $[10^{-4}, 5 \times 10^{-2}]$ is also applied. We use RMSProp with an initial learning rate log-uniformly sampled from $[10^{-5}, 5 \times 10^{-2}]$.

# B Additional Experimental Results

In Section 4, we tested *random starts* and *sticky actions* as either regularizers during training or evaluation strategies during testing. Figure 5 and Figure 6 shows additional experiment results for the regularization tests on the `BLOCKS8` and `TUNNEL` mazes, respectively. As we can see, a combination of both regularizers makes it harder for the agents to fit a large number of random mazes in the `BLOCKS8` environments. However, on the harder `TUNNEL` mazes, the training scores are not significantly different from the results without regularizations. The regularizations do improve the test scores. Nevertheless, serious overfittings can still be observed.

Figure 7 presents complete studies for different evaluation strategies. As also mentioned in the main text, when the training set is small, or for the case of random mazes, the performances evaluated on the test set are very different from the performances evaluated on the training set with additional stochasticity. With "large" enough training set and "small" enough random-reward noises, the evaluation results could be close under all protocols. However, specific values for "large" and "small" heavily depend on the underlying tasks. For example, for `BASIC` mazes with no noises, 100 training levels is enough to make the evaluation performances similar, but for `TUNNEL` mazes, more than 1,000 training levels are needed.

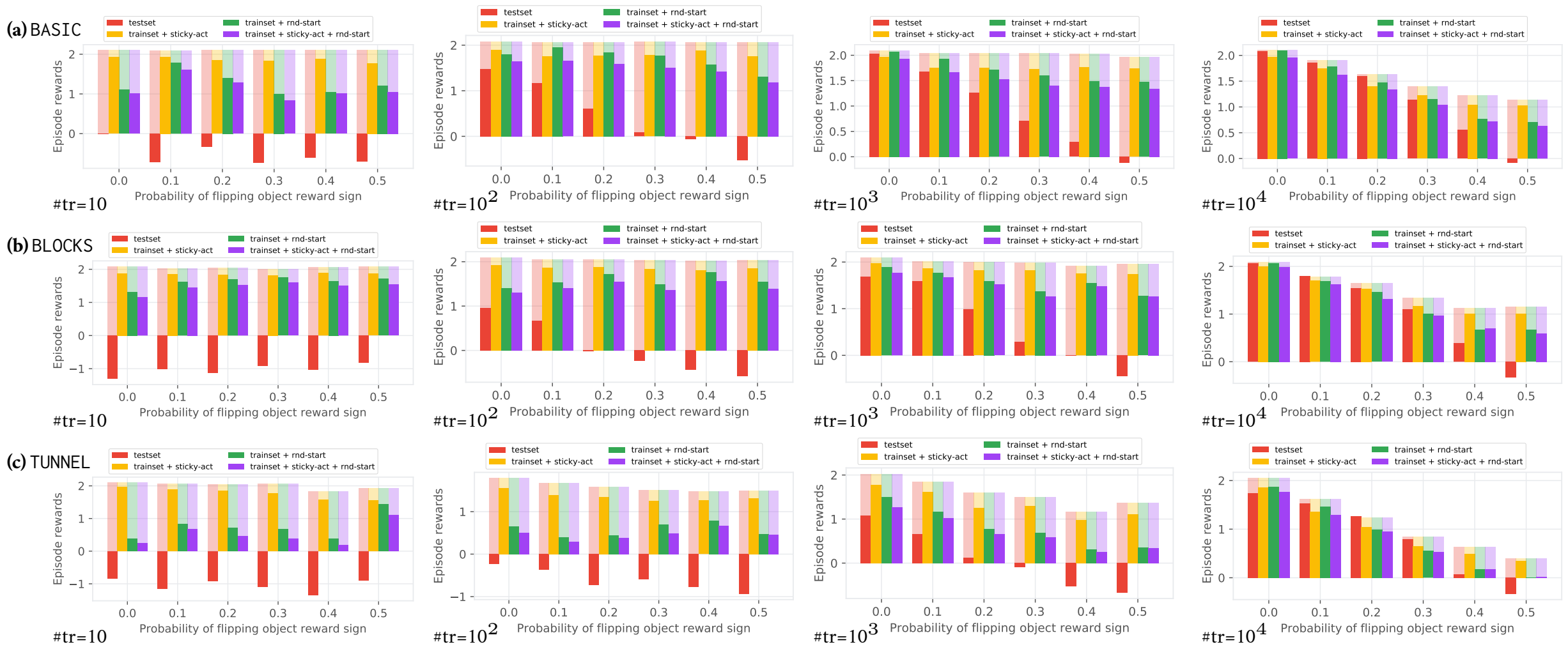

Figure 7: Evaluation of trained agents with different protocols. The rows are results on (a) BASIC, (b) BLOCKS and (c) TUNNEL mazes, respectively. The columns index agents trained with 10, 100, 1,000 and 10,000 mazes, respectively. Each plot is formatted in the same way as Figure 3b.

# C Analysis of Agent Trajectories

In this section, by visualizing the trajectories of trained agents in various evaluation conditions, we provide qualitative analysis[5] on different kinds of overfittings and how some standard evaluation techniques could fail to detect them. In particular, we show that agents trained with small training set could overfit by memorizing a "soft" lookup table for solving the training set levels without understanding the dynamics of the maze. By "soft" we mean that the memorization is robust to small perturbations, therefore, stochastic policies and sticky actions cannot detect it. But random starts that put agents in a remote starting point (causing big changes to the states) could. On the other hand, when trained with a large training set with random rewards, the agents still overfit (by construction) but with proper understanding of the underlying environment dynamics. In this case, random starts will also fail to detect overfitting effectively. Note all the agents analyzed in this section are trained *without* random starts or sticky action regularizations.

## C.1 Stochastic Policies

In this section, we demonstrate a basic example of how a trained agent could generate different trajectories with small variations. This example also serves the purpose of introducing the notations in our trajectory visualizations. In particular, Figure 8 shows a full trajectory of a level on the *training set*, generated by an agent trained with only 10 training levels. We call this agent Agent-1.

In the visualization, an episode is broken into several pictures containing segments of the full trajectory, mainly to avoid clutters (e.g. when an agent steps back into the previous location). In each picture, black squares indicate obstacles and walls, circles indicate consumable objects. Colored squares indicate the location of the agent, where the color from light to dark indicate the time ordering in a segment of the trajectory. Within each square, a symbol is draw to indicate the proposed action by the agent when in that location: arrows are moving directions, and a cross sign means "stay". In each segment, the trajectory from the previous segment is also draw with light gray to help visualization. At the top of each maze visualization, text shows the frame numbers and the current cumulative rewards at the end of this segment. For technical reason, the final frame, *after* the agent consumes the terminating-object (or *after* the environment times out), is not shown. But that does not affect our analysis of the trajectories.

[5]Note all the conclusions are based mainly on the quantitative results presented in the paper. The visualizations should be considered supplementary. In particular, although on average the behaviors and performances show patterns, one particular trajectory rollout could be good or bad by chance due to the stochasticity. We strictly follow the evaluation protocols, but we do cherry-pick specific mazes that we believe are representative for each situation.

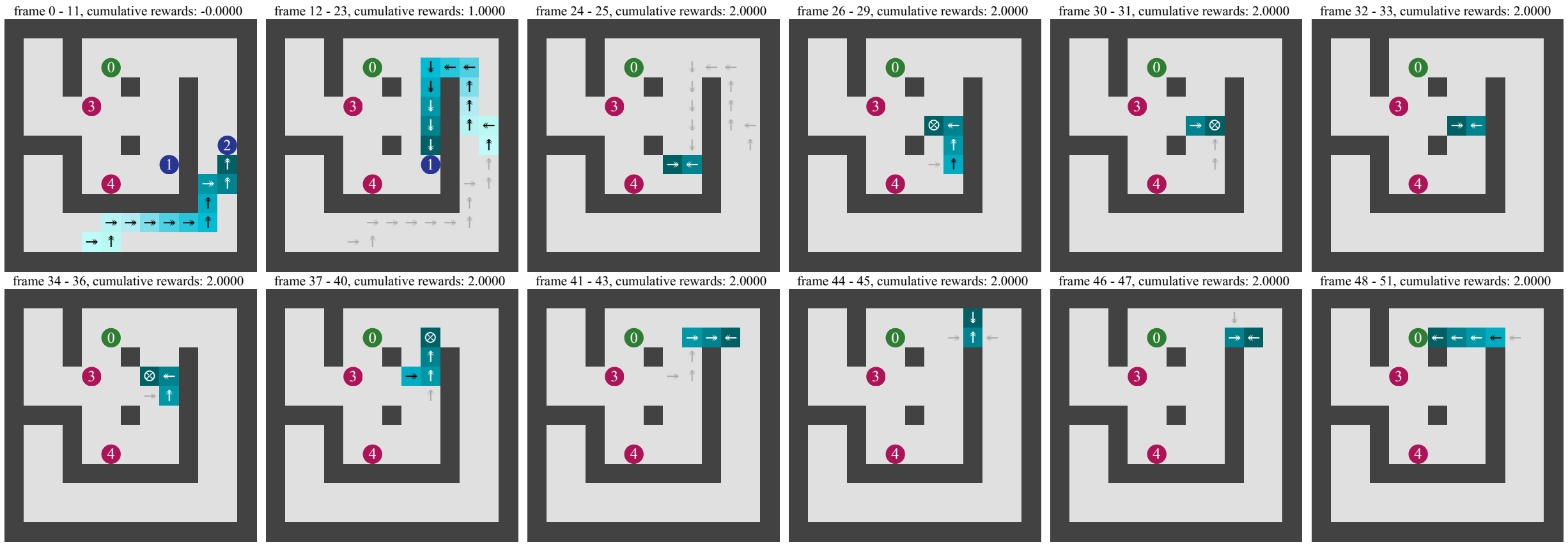

Figure 8: Trajectory visualization of an episode for a level in the training set, for an agent (`Agent-1`) trained with 10 training levels on the `TUNNEL` maze environment. The trajectory is broken into segments to avoid overlappings in agent locations.

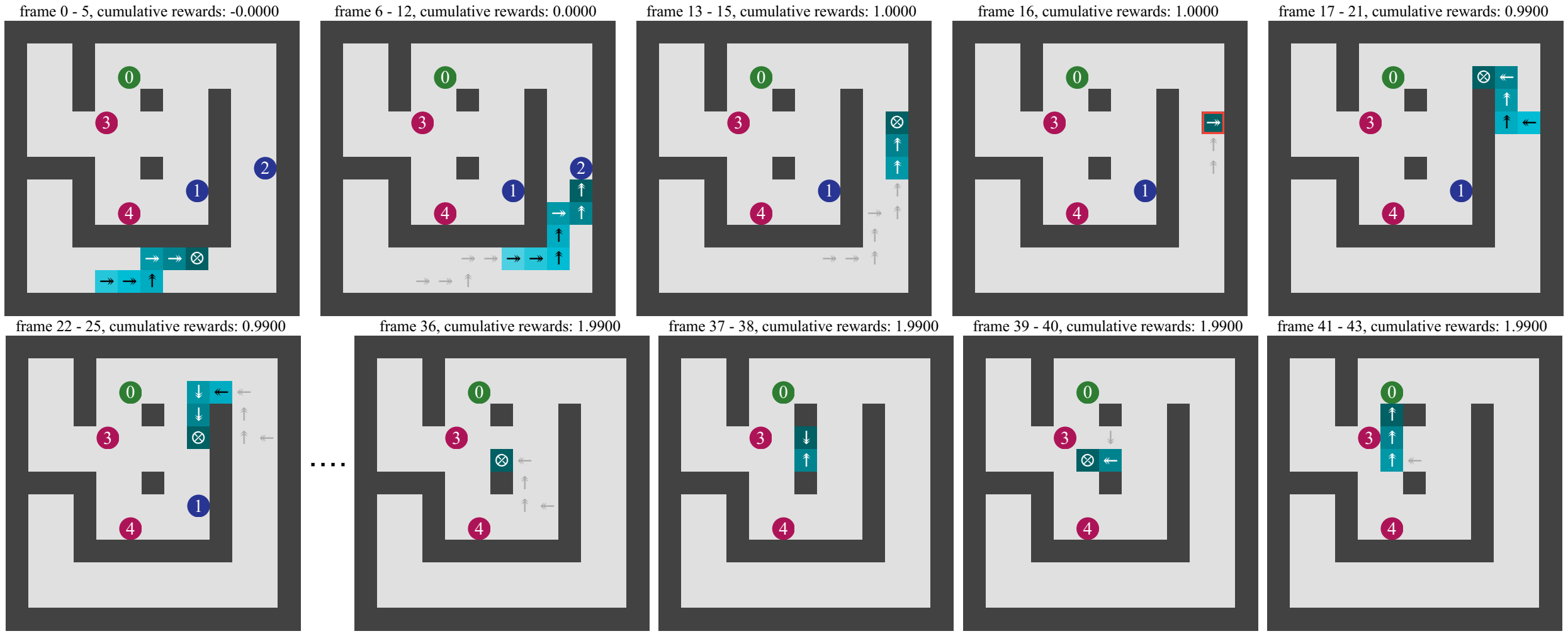

Figure 9: Trajectory visualization of an episode for a level in the training set, for an agent (`Agent-1`) trained with 10 training levels on the `TUNNEL` maze environment. Note for brevity we skipped frame 26 $\sim$ 35. This is a different sample of trajectory by the same agent (`Agent-1`) on the same maze as in Figure 8. The routes in both cases are similar but with small variations.

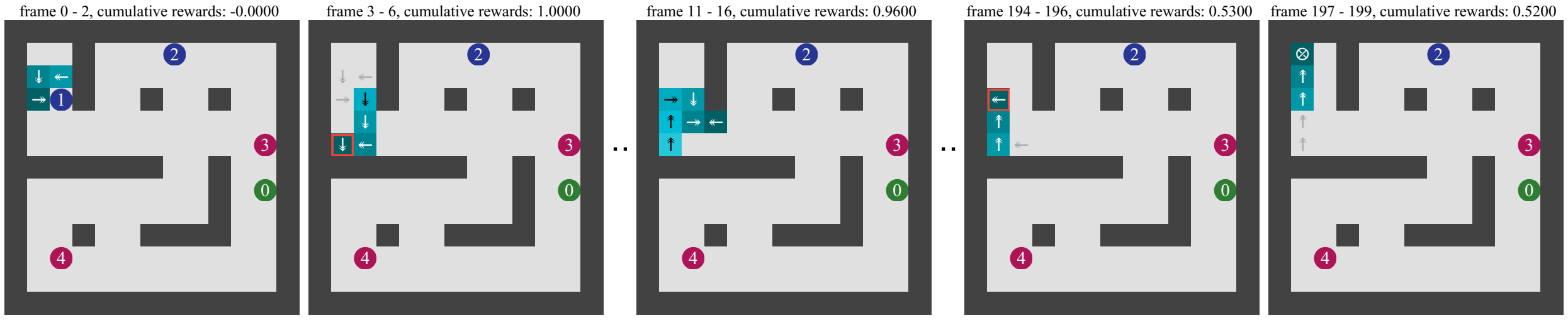

Figure 10: Trajectory visualization of of an episode for a level in the test set, for an agent (`Agent-1`) trained with 10 training levels. The full trajectory is very long, therefore we only show part of it. The agent starts out in the top left corner, consumes the nearby object 1, but has never been able to "escape" that corner until frame 199, when the episode ends with a timeout penalty.

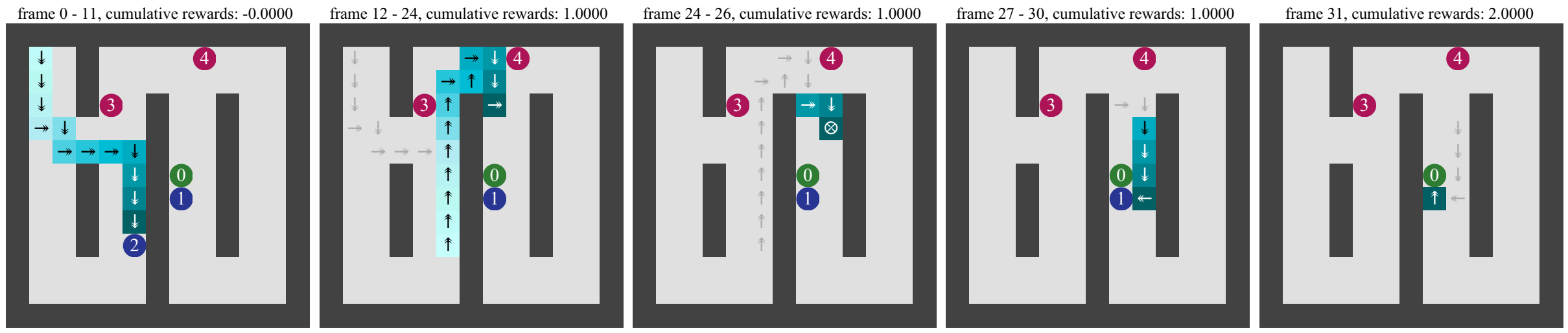

Figure 11: Trajectory visualization of an episode for a level in the training set, for an agent (`Agent-1`) trained with 10 training levels.

As shown in Figure 8, `Agent-1` goes straight to consume object 2 and then 1. Afterwards, it wander around before finally consuming object 0 to terminate the episode. As mentioned before, the agent propose an action based on a multinomial distribution over possible actions calculated by the current policy. In Figure 9 we show an alternative trajectory generated by the same agent on the same maze. The route is similar but with small variations. Note on frame 16 (the 4th picture in Figure 9), the agent attempts to walk into the wall, we mark an action that has failed to execute with a red bounding box. This will also be useful when we analyze the behaviors of *sticky actions*.

## C.2 Behaviors of an Overfitted Agent

In this section, we look at the behavior of `Agent-1` under different evaluation protocols. The quantitative evaluations in Section 3.2 have shown that agents trained with only 10 levels do not perform well on unseen test levels. Figure 10 shows an example trajectory of `Agent-1` on an unseen test maze. It struggles around the initial location until timeout after 200 steps. Although the full trajectory is not shown, from the final cumulative rewards, we can infer that the agent gets a lot of penalty attempting to walk into walls. It clearly does not demonstrating any hint of "understandings" of how a maze works.

In Figure 11, we visualize the trajectory taken by `Agent-1` on a maze level in the *training set*. This time, the agent behavior near optimally and achieves the maximum rewards efficiently. In Figure 12 and Figure 13, we visualize the agent behaviors on the same maze (from the *training set*), but with added sticky action (stickiness 0.25) and random starts to the evaluation environments.

For sticky actions, the agent still achieves the optimal rewards, although it takes a bit longer time. For example, in the 3rd picture of Figure 12, the agent tried to move to the right, but the environment executed the previous action and moved the agent one step up. The agent then corrected this by moving down, and proposing to move right again (as shown in the 4th picture). This time, it was unlucky again and the environment executed the previous action and moved it downwards. In summary, the sticky actions create random "local" perturbations to the states, but the agent is able to deal with those robustly even though it is not explicitly trained with sticky actions.

On the other hand, in Figure 13, even though the maze is exactly the same, when started at a different initial location, the agent struggled through the whole episode and never accomplished anything.

In summary, `Agent-1` overfits to the 10 training levels without developing a good "understanding" of how mazes work. It has troubles dealing with unseen mazes or larger modifications like random starts during evaluation. However, it handles sticky actions reasonably well. This is potentially because sticky actions only create small "local" perturbations to the states, and it is likely that the agent has seen the same or very similar states in its own stochastic rollouts during training. Unlike algorithms like *Brute* (Machado et al., 2017) that optimize over open-loop sequences of actions without even looking at the states, agents trained with neural network approximators could potentially overfit while still being robust to small perturbations.

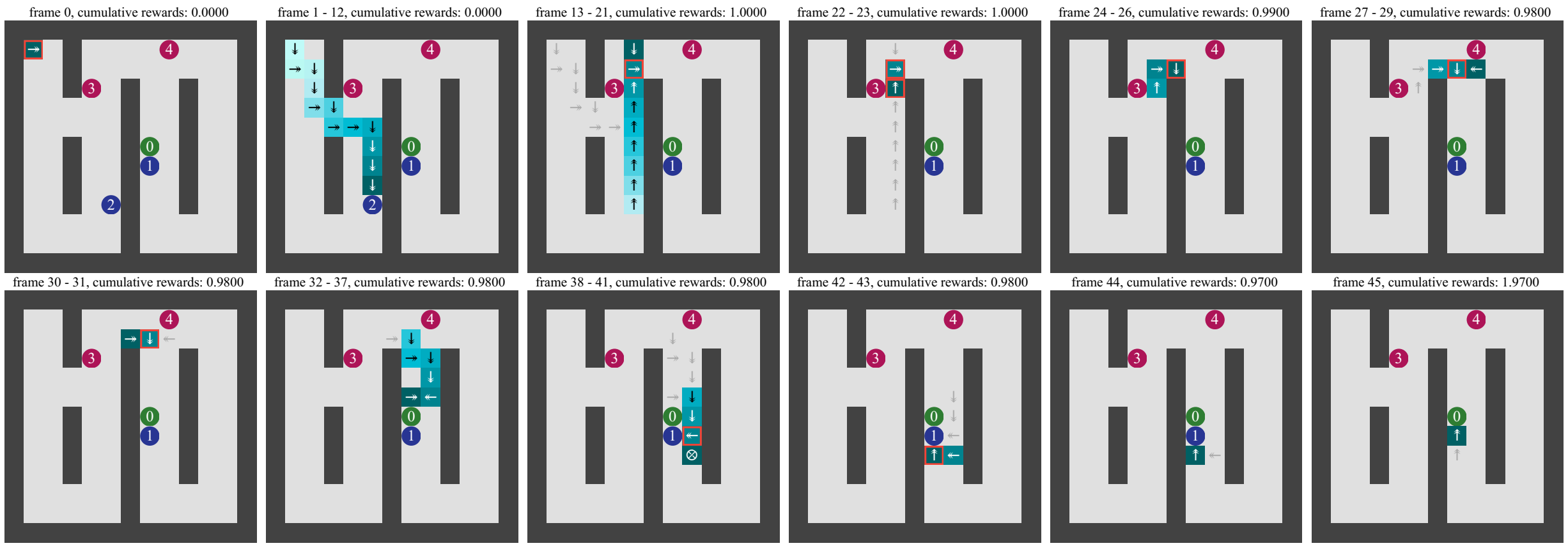

Figure 12: Trajectory visualization of an episode for a level in the training set with added action stickiness of 0.25 during evaluation, for an agent (`Agent-1`) trained with 10 training levels. The training is done *without* sticky actions. The red bounding box indicate an action that is unsuccessful, mostly due to action stickiness of the environment, and sometimes also due to invalid actions like trying to walk into the wall. For example, in the first picture, the agent tries to move right, but as we can see from the second picture, it ended staying, due to the previous action (default is "stay" at the beginning of an episode) being repeated. From the second picture, the agent chooses to move downward instead — showing that the agent itself has a stochastic policy. In the 3rd picture, for the second to the last action, the agent wanted to go right, but ended up one block above due to the previous action (going up) gets repeated. This is the same agent (`Agent-1`) and level as in Figure 11. Although it takes longer, the agent successfully collects all positive-rewarded objects and achieved the optimal rewards, minus small penalties due to some invalid action attempts.

**Behaviors of a Well-trained Agent** For comparison, we visualize the trajectories on the same maze of a well trained agent, `Agent-2`, trained with 10,000 training levels. In particular, Figure 14 and Figure 15 show the trajectories with added sticky actions (stickiness 0.25) and random starts, respectively.

Figure 14 is an example of the unlucky cases for sticky actions, where an agent accidentally run into a negative-reward object (object 3 in this case) due to action stickiness. Although the agent finished the rest of the episode optimally, the final reward was still significantly below the optimal. However, we expect this kind of unlucky cases to happen with roughly equal probability to all agents, therefore comparing of *relative* scores among agents is still fair.

Figure 15 shows two different examples of random start locations. The left hand side shows exactly the same initial configurations that caused `Agent-1` to get stuck until timeout (see Figure 13). For `Agent-2`, it handled this situation effortlessly, although it somehow chooses to skip object 2. Note there is no moving cost in the maze environment, so in principle it is better to take object 2, even though it is far away. While the strategy taken by `Agent-2` may be suboptimal, it clearly demonstrates sensible skills navigating the maze (comparing to `Agent-1` in Figure 13). The right hand side of Figure 15 shows a different realization of random starts, when the agent was spawn at the lower left corner. This time it took all positive-reward objects and achieved the optimal rewards.

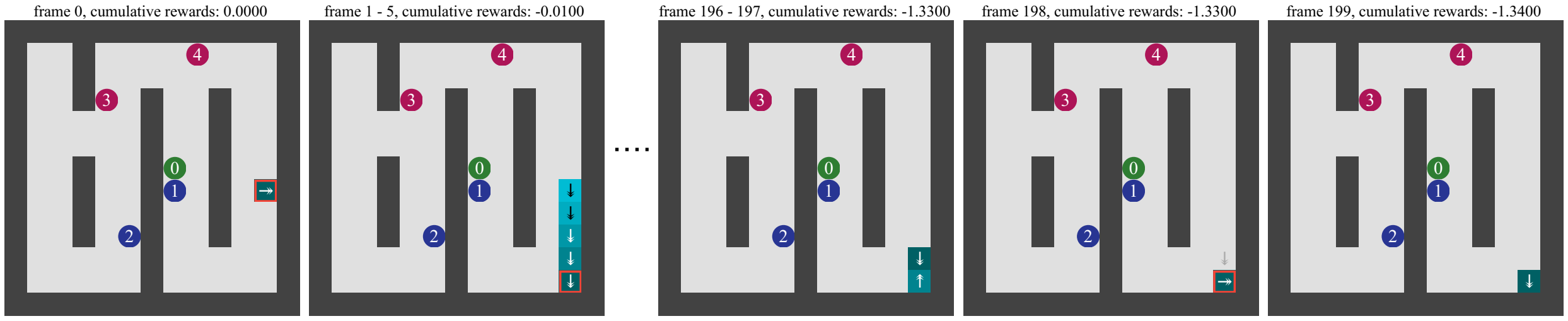

Figure 13: Trajectory visualization of an episode for a level in the training set with added random starts during evaluation, for an agent trained with 10 training levels. The training is done *without* random starts. This is the same agent and level as in Figure 11: the agent initial location is different due to random starts. The agent is "trapped" in the bottom right corner for the whole episode until timeout after 200 steps. For brevity, only part of the trajectory is shown. The final cumulative rewards also indicate that it gets a lot of invalid-action penalty attempting to walk into walls.

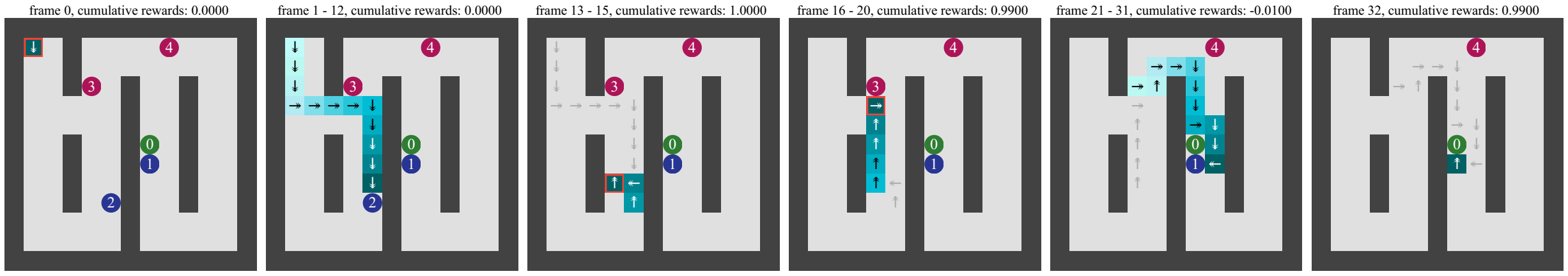

Figure 14: Trajectory visualization of an episode for a level in the training set with added sticky actions (stickiness 0.25) during evaluation, for an agent (`Agent-2`) trained with 10,000 training levels. The training is done *without* sticky actions. This is the same maze level as in Figure 11. In the 4th picture, the action stickiness accidentally pushed the agent into a negative-reward object (object 3).

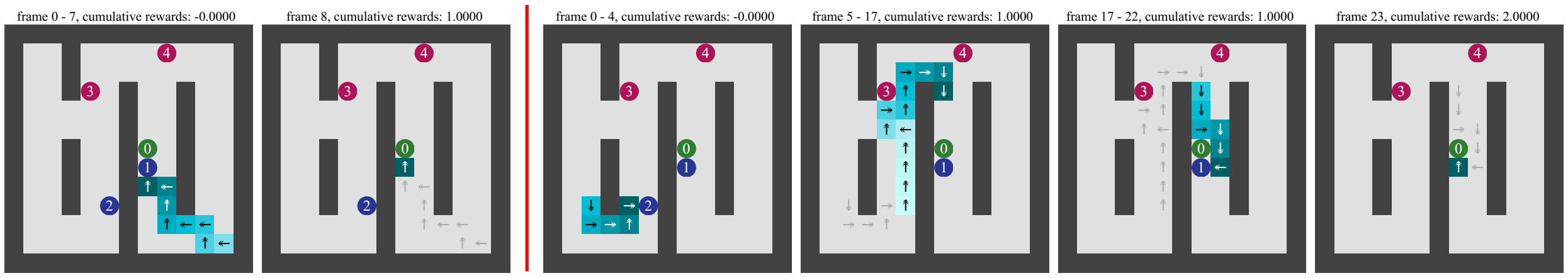

Figure 15: Visualization of two trajectories for a level in the training set with added random starts during evaluation, for an agent (`Agent-2`) trained with 10,000 training levels. The training is done *without* random starts. The maze level is the same as in Figure 11. The two trajectories show two different random start locations for the agent. Also compare with Figure 13 for the behavior under random starts of an overfitted agent (`Agent-1`).

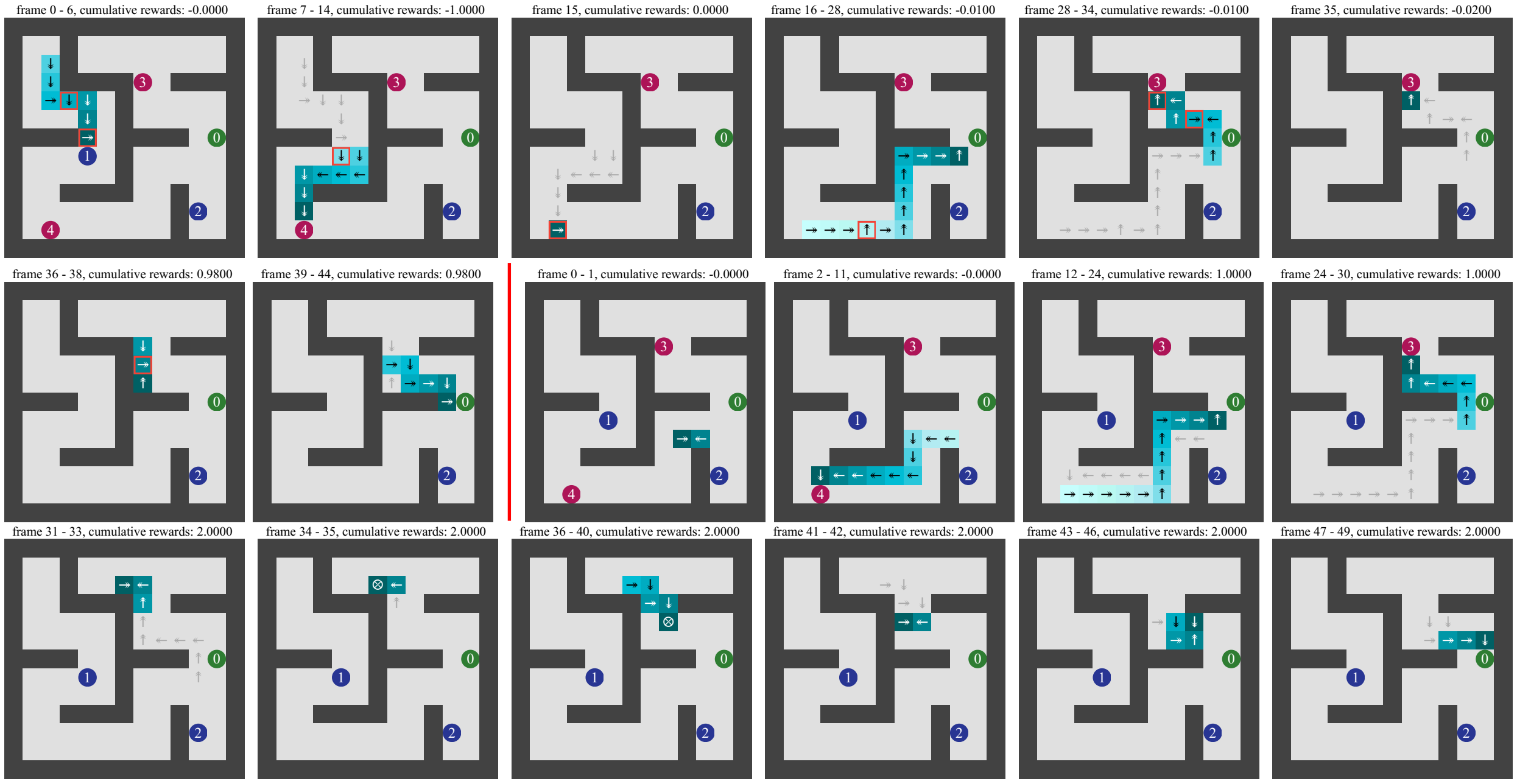

Figure 16: Visualization of two trajectories for a level in the training set with added sticky action (stickiness 0.25) and random starts, respectively. The agent (`Agent-3`) is trained with 1,000 training levels with random rewards. For this particular maze level, object 3 and 4 are of reward 1, while object 1 and 2 are of reward -1.

### C.3 An Overfitted Agent that Knows about Navigation

In this section, we look at an agent, called `Agent-3`, trained on 1,000 training levels with pure random rewards. Note by construction, the only thing that `Agent-3` can do is to memorize the training mazes. Because there is no statistical patterns to learn, it performs badly on unseen test mazes, as verified in the previous quantitative results in this paper. We also know that those agents achieves superficially high scores when evaluated on the training set with sticky actions or random starts. In Figure 16, we visualize two trajectories generated by this agent on a training maze level with added sticky action (stickiness 0.25) and random starts, respectively. In this particular maze level, due to reward flipping, object 3 and 4 are of rewards +1, while object 1 and 2 are of rewards -1. From the visualization of the trajectories, we can see that `Agent-3` understands this situation and navigates the maze without efforts in both cases. Apart from the unlucky situation of accidentally stepping into object 1 via sticky action (see the 1st picture of Figure 16) thus incurring a penalty of -1, the agent performs optimally.

The visualization of the trajectories demonstrates that, despite that it completely overfits the training set with random rewards, `Agent-3` learns how to navigate a maze after consuming a large training set. In this case, it can handle even large perturbations like random starts.

### C.4 Sticky Action vs Random Starts

From the previous studies, we can see that for our maze environment, random starts add larger perturbations to the evaluation procedures than sticky actions. As a result, while it is relatively easy for the agents to perform well under sticky actions, handling random starts requires the agents to really learn to navigate the maze reasonably well. Moreover, sticky actions could lead to unlucky situations where an agent is accidentally moved into a negative-reward object, introducing bias to the final rewards that is hard to characterize or estimate. Therefore, random starts is better than sticky actions for evaluation in our mazes, although random starts still fail to detect overfitting in the case of random reward mazes.

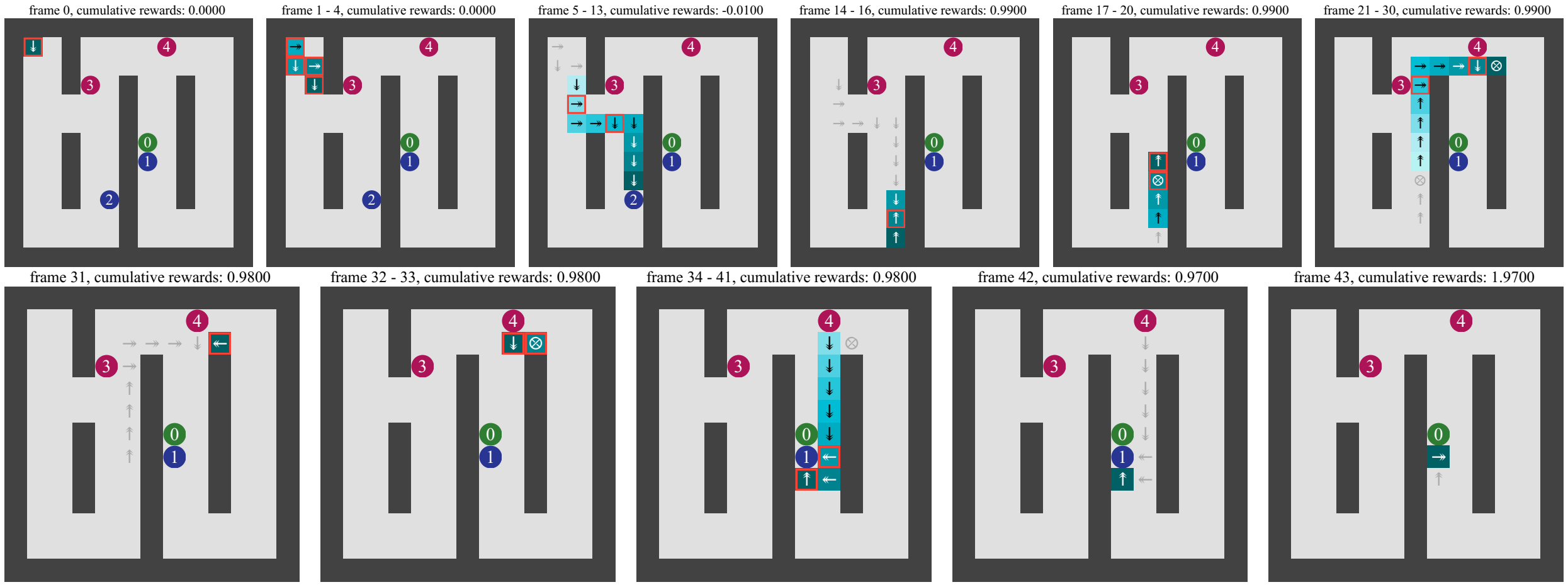

Figure 17: Trajectory visualization of an episode for a level in the training set with added action stickiness of 0.99 during evaluation, for an agent (`Agent-1`) trained with 10 training levels. The training is done *without* sticky actions. The maze level and agent is the same as in Figure 11. For stickiness 0.99, almost all the actions executed by the environment are the previous one. In the visualization, some proposed actions are not marked with red bounding box because they are successfully executed since the previous action and the current one are the same. In this particular case, the agent managed to achieve the optimal rewards despite strong action stickiness in the evaluation process.

On the other hand, Machado et al. (2017) suggested to use sticky actions over random starts for evaluating Atari games. One of the main reasons is the difficulty to arbitrarily control the games in the Atari simulators as they are not originally built for RL purpose. Therefore, it is hard to generate random and valid start states for many games. Alternative approaches include starting from the middle of a human replays or initial sequences of no-ops, but those workarounds inevitably introduce biases in the sampled random starts.

In summary, there are trade-offs when using either random starts or sticky actions as evaluation strategies. The effectiveness critically depends on the nature of each specific task. In general, using held-out test sets is a more direct evaluation than applying those strategies to the training sets.

# D Different Levels and Types of Action Stickiness

During our studies, we also tested the sticky actions with different levels of stickiness. Surprisingly, we found that the agents are robust to even very large values of stickiness in our maze environments. For example, Figure 17 visualizes a trajectory generated by `Agent-1` studied in Subsection C.2, on the same training level as in Figure 11, but with added sticky actions of the stickiness 0.99 during evaluation. With such high values of stickiness, almost all the proposed actions by the agent would be ignored. The environment will choose to always execute the action proposed in the previous frame instead. Note not all action proposals are marked by red border in Figure 17 because in some cases the previously proposed action and the currently proposed action are the same, therefore the currently proposed action is successfully executed by luck.

Despite that most of the proposed actions are not executed properly, and the agent has not experienced sticky actions during training, it is still able to reach to optimal rewards relatively easily. After inspecting this and some other trajectories, we argue a reasonable explanation for the robustness is that the states of the current frame and the previous frame are semantically close most of the time. Therefore, executing an action proposed either for the previous frame or for the current frame could both bring the agent closer to the goal.

An alternative way of defining sticky actions could be to run (with probability given by the stickiness value)

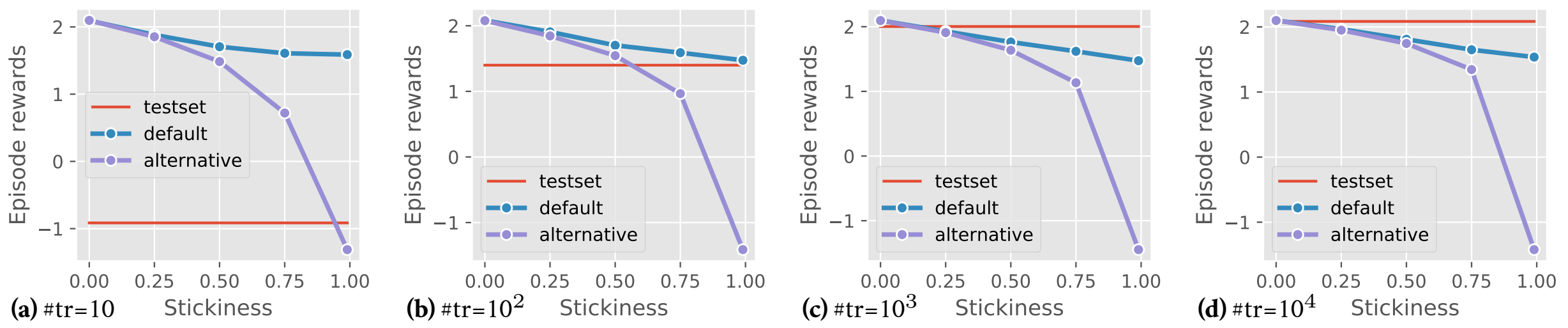

Figure 18: Evaluation with different levels and different modes of action stickiness on the *training set*, for agents trained with the number of training levels being (a) 10 (b) 100 (c) 1,000 and (d) 10,000. The agents are trained *without* sticky actions, on the BASIC mazes, and without random flipping of reward signs. The "default" curves show the evaluation scores tested on the training set with the *default* mode of sticky actions defined in (1), and the "alternative" curves show the evaluation scores tested on the training set with the *alternative* mode of sticky actions defined in (2). The red line shows the evaluation performance on the unseen test set for reference.

the action *executed* in the previous frame, as oppose to the action *proposed* in the previous frame. In this mode, stale actions could be carried over multiple frames, affecting the behavior in a state that is (semantically) further away. More specifically, with a *stickiness* parameter $\zeta$, at time $t$, the action $A_t$ that the environment executes is

$$A_t = \begin{cases} a_t, & \text{with prob. } 1-\zeta \\ a_{t-1}, & \text{with prob. } \zeta \end{cases} \tag{1}$$

for the *default* mode of sticky actions as defined in Machado et al. (2017, Section 5.2). In the *alternative* mode, it is

$$\tilde{A}_t = \begin{cases} a_t, & \text{with prob. } 1-\zeta \\ A_{t-1}, & \text{with prob. } \zeta \end{cases} \tag{2}$$

For the *alternative* mode, with stickiness 0.99, *all* agents will fail the evaluation because the environment will be only executing the initial default action ("stay") almost all the time. In Figure 18 we plot the performance evaluations with different modes and different level of action stickiness. As we can see, for the *default* mode, the evaluation scores drops only by a small amount, but for the *alternative* mode, the evaluation scores drops to negative with high values of stickiness. However, this behavior is indiscriminative with respect to the actual generalization performance of the agents. For reference, in each plot, a red line is drawn to show the performance evaluated on the held-out test set. As we can see, with increasing number of training levels, the performance on the unseen mazes increases. However, the curves for the scores evaluated with both modes of sticky actions remains almost identical. Therefore, in this case, all of the combinations of stickiness levels and modes remain ineffective at distinguishing the generalization power of trained agents.